\documentclass[letterpaper]{article} 
\usepackage{aaai2027}  
\usepackage[hyphens]{url}  
\usepackage{graphicx} 
\usepackage{natbib}  
\usepackage{caption} 
\usepackage{algorithm}
\usepackage{algorithmic}
\usepackage{amsmath}
\usepackage{amssymb}
\usepackage{bm}
\usepackage{booktabs}
\usepackage{multirow}

\usepackage{newfloat}
\usepackage{listings}
\DeclareCaptionStyle{ruled}{labelfont=normalfont,labelsep=colon,strut=off} 
\floatstyle{ruled}
\newfloat{listing}{tb}{lst}{}
\floatname{listing}{Listing}

\usepackage{booktabs}

\newcommand{\E}{\mathbb{E}}

\definecolor{suppblue}{rgb}{0.00,0.28,0.72}
\newif\ifsuppcolor\suppcolortrue

\newcommand{\best}[1]{#1}
\newcommand{\sbest}[1]{#1}

\title{Unsure but Certain: Uncovering the Representation-Confidence Gap in Diffusion Language Models}
\author{
    Saurabh Yadav,
    Badri Narayana Patro,
    Vijay Srinivas Agneeswaran
}
\affiliations{
    \textsuperscript{\rm 1}Microsoft\\

    India\\
}

\begin{document}

\maketitle

\begin{abstract}
Diffusion language models use broad context to create text, suggesting they might handle input noise better than standard models. Testing reveals this is only partially true. Internally, diffusion models detect text errors highly accurately. Externally, their reported certainty ignores this signal. As accuracy drops due to noise, confidence stays near its maximum and the ability to correctly rank answers degrades toward random chance. We call this mismatch the representation confidence gap. The visible concentration of high certainty scores is a misleading surface symptom. Standard math adjustments remove this concentration but fail to fix the underlying loss of ranking order. This ranking deficit favors standard models under noisy conditions and resists common remedies. Matching training recovers accuracy but not ranking, while score recalibration and input level error signals cannot reorder the final answers. However, the information needed to properly evaluate an answer survives in the hidden states. A lightweight extraction tool uses this signal to improve ranking. This approach is highly efficient because it leaves the base model completely frozen and requires zero additional text generation steps. We present this tool to prove the signal exists, while clearly noting its limits. Ultimately, certainty reliability is a more pressing limit than overall accuracy under noisy conditions.
\end{abstract}


\section{Introduction}
\label{sec:intro}

Diffusion language models like LLaDA \citep{llada2025} and DREAM \citep{dream2025} create text by revealing a full sequence with access to broad context instead of guessing one word at a time \citep{d3pm2021,mdlm2024,sedd2024}. This broad view suggests they might handle input noise better than standard models. Testing reveals this assumption is only partially true against the noise real systems actually encounter.

\begin{figure*}[t]
\centering
\includegraphics[width=\textwidth]{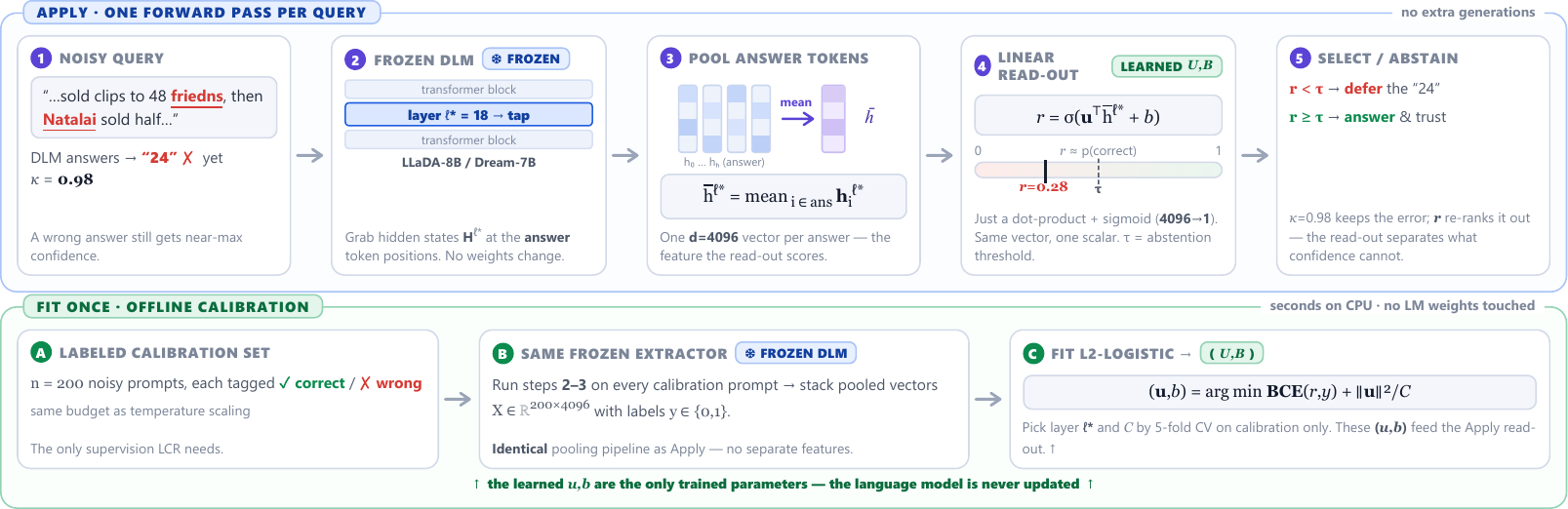}
\caption{The representation confidence gap and the Latent Correctness Readout. Under noisy conditions the model answers incorrectly while still reporting a high certainty score, so certainty no longer clearly separates correct answers from incorrect ones. \emph{Apply} (top): in one single pass we average the answer word numbers at a specific layer, score them with a simple mathematical tool, and skip the question if the score falls below a set cutoff. \emph{Fit once} (bottom): the simple tool is trained offline on a small set of labeled examples using the same frozen extraction process, leaving the base model weights completely untouched.}
\label{fig:overview}
\end{figure*}

We tested LLaDA 8B and DREAM 7B on logic puzzles using four common types of text errors to measure accuracy and certainty reliability. The internal states of a diffusion model detect errors highly accurately. However, the final output ignores this signal. As accuracy drops due to noise, overall certainty remains high and the ability to correctly rank answers degrades toward random chance. Adjusting the scale of these scores fails to fix the problem because it does not change the order of the answers. The model frequently predicts the wrong answer while reporting high certainty. We call this mismatch the \emph{representation confidence gap} (Figure~\ref{fig:overview}).

Previous studies show that internal states of a standard model track correctness better than final probabilities \citep{burns2022discovering,azaria2023internal,kossen2024semantic}. We focus on how much of this internal signal survives in the final score when the input text has errors. This signal loss is notably larger for diffusion models. Under noisy conditions, standard models retain more ranking ability than diffusion models in our specific tests. On clean text, the results are much closer across the groups.

Diffusion certainty is heavily concentrated near the very top of the scale. For example, LLaDA reports high certainty on the vast majority of its mistakes when facing errors, while the standard LLaMA 3 8B model rarely does this. This visible concentration is a misleading surface symptom. A standard temperature adjustment lowers the overall certainty scores without changing the actual ranking order. When evaluated at this adjusted point, the large difference between the model types shrinks significantly. The concentrated high scores mask the underlying loss of proper ranking. This shows that the way diffusion models sort their guesses does not entirely cause the main problem.

We separate accuracy and certainty to evaluate robustness. Accuracy is straightforward to recover. A basic matching training method recovers most of the lost accuracy, performing as well as three specific diffusion methods we tested. However, this training does not fix the certainty problem. It simply forces noisy guesses to match clean guesses, which were already highly confident. Standard math adjustments \citep{li2026confidence} also fall short for a structural reason. They reduce the overall error measurement but keep the answers in the exact same order. Because the order stays the same, the model still struggles to safely skip uncertain questions. Noise degrades the ranking, and simple score adjustments cannot restore it.

A useful signal to improve this ranking is present inside the model. Our error detector looks at the whole question, so its score is mostly the same across all possible answers. Adding this score changes the certainty of the whole question without reordering the answers, acting just like basic math adjustments. Two methods that try to feed this signal into the final score fail to improve ranking exactly as expected. Fixing the ability to skip questions requires a signal linked to the specific answer. Therefore, we introduce the \textbf{Latent Correctness Readout}, or LCR. This is a lightweight extraction tool that looks at the internal states of the answer words to estimate correctness. This approach is highly efficient because it leaves the base model completely frozen and requires zero additional text generation steps. It fits quickly using the same small set of examples needed for basic adjustments.

Our contributions are as follows. First, we define the representation confidence gap in diffusion language models under real world text noise. We separate the surface issue of concentrated high certainty scores from the deeper problem of lost ranking ability. Second, we explain the structural reasons why obvious fixes fall short. Matching training and basic score adjustments keep the answers in their original order. Furthermore, question level error signals stay mostly the same across all possible answers, so they cannot reorder the final answers either. We support this claim by showing two routing methods that did not improve results. Third, we show that the information needed to skip questions safely survives in the hidden states. We extract this signal using a lightweight tool that is highly efficient, requiring no model updates and zero additional generation passes. We present this tool to prove the signal exists, while clearly noting its limits.
\section{Background and Problem Setup}
\label{sec:bg}

\textbf{Masked diffusion language models}
A masked diffusion model generates text by reversing a process that hides it \citep{d3pm2021,mdlm2024}. During generation, the model runs for a set number of steps. At each step, every hidden position receives a probability distribution. The system commits to positions with the highest probabilities and leaves the rest for later. This certainty based scheduling is unique to diffusion generation. We record the certainty for each answer position at the step it is revealed.

\textbf{Confidence calibration and selective prediction}
\label{sec:calib-setup}
We score a generated answer using the geometric average of its individual word certainties
\begin{equation}
\label{eq:conf}
c(\bm{q})=\Big(\textstyle\prod_{i\in\mathcal{A}}\kappa_i^\star\Big)^{1/|\mathcal{A}|},
\end{equation}
taken over the answer positions. This average matches the model's training math and is heavily influenced by the least certain word. It includes a selection effect because the system only records scores when they are the highest among the remaining hidden words. This selection changes the overall certainty scale, which we later show is separate from how the model ranks answers. We evaluate this score in two ways. Calibration checks if certainty matches actual accuracy using expected calibration error \citep{naeini2015,guo2017}. Selective prediction \citep{geifman2017} checks if the certainty score helps the model skip questions, balancing answer rates against error risks. We measure this ranking quality using the area under the receiver operating characteristic curve, where 0.5 is random chance. Distinguishing calibration from ranking is crucial because simple math adjustments can fix calibration without reordering the answers. Correcting the certainty scale is completely different from fixing answer ranking.


\section{Diagnostic Findings on Model Robustness}
\label{sec:diag}

We summarize the findings motivating our approach here, detailing the full setup in the Experimental Setup section. Unless stated otherwise, numbers apply to LLaDA 8B on the GSM8K test under character swapping noise. Every clean versus noisy comparison uses the exact same 400 prompts for fairness.

\textbf{Detecting errors from internal states:}
A simple tool evaluating frozen internal states separates clean from noisy text with very high reliability, scoring between 0.996 and 0.997 across multiple layers. A restricted version of this detector still reaches 0.986. The model clearly registers when its input contains errors.

\textbf{Certainty remains high under noise:}
While accuracy drops from 0.560 to 0.373 under noise, the average certainty score barely moves, falling from 0.988 to 0.982. Meanwhile, calibration error rises from 0.428 to 0.610. These models are already highly confident on clean text, and noise worsens this by lowering accuracy while leaving high certainty mostly untouched.

\textbf{Certainty loses connection to accuracy:}
This failure goes beyond a simple scale shift. On the noisy test set, certainty is higher for correct answers than incorrect ones by only 0.0022. This tiny gap of two parts per thousand is useless for deciding whether to trust a specific answer, especially given the much larger accuracy gap. Ranking quality drops from 0.666 to 0.575. We describe this behavior as being unsure of the answer but certain in the reported score.

\textbf{Standard fixes do not transfer:}
Standard temperature adjustments \citep{li2026confidence} cannot repair a broken ranking order. On clean text, they reduce calibration error significantly while leaving useful ranking untouched. Under noise, they reduce calibration error from 0.610 to 0.133 but leave the ranking score frozen at 0.575. The adjusted temperature reaches the absolute limit of our search range, flattening the certainty scores completely. Expanding this search range would only change calibration error, not the underlying ranking failure.

Basic matching training recovers most accuracy lost to noise, and three special diffusion methods we tested added no reliable improvement. Accuracy under noise is manageable with existing tools, but certainty reliability requires a different approach.

\section{Method: From Corruption Routing to a Correctness Readout}
\label{sec:method}

We build on a basic training method that restores accuracy and explore how to obtain a reliable certainty score. We initially tried to feed the model's internal error signal into its final certainty output. This attempt fails for structural reasons, pointing to the need for a scoring method focused specifically on the generated answer.

\textbf{Accuracy baseline and consistency}
\label{sec:cons}
We train the model to match its noisy predictions to its clean predictions
\begin{equation}
\label{eq:cons}
\mathcal{L}_{\mathrm{cons}}(\theta) = \E_{i \in \mathcal{A}} \left[ \text{KL}\left( \text{sg}\left[f_{\theta_0}(\cdot \mid \bm{q})_i\right] \parallel f_{\theta}(\cdot \mid \tilde{\bm{q}})_i \right) \right],
\end{equation}
where $\theta_0$ is the frozen base model and $\text{sg}[\cdot]$ stops the gradient update. This approach recovers most of the lost accuracy. However, because it copies the clean text behavior it also copies the original overconfidence, leaving the certainty problem unfixed.

\textbf{Routing the error signal and why it fails:}
\label{sec:routing}
From the error detector we obtain a frozen score $g_i=\sigma(\beta(\bm{w}^\top \bm{h}_i^{\ell^\star}(\tilde{\bm{q}}) - b))\in[0,1]$ at each word position. We tested two methods to feed this score into the final certainty: a training penalty that lowers certainty at flagged positions and a decoding rule that delays decisions when errors are detected. Neither method improved ranking over the basic training. While the training penalty achieves the highest accuracy we measured, its ranking ability shows no statistical improvement.

This failure happens for a structural reason. Skipping questions successfully depends on the relative order of certainty scores for different possible answers. The error detector score is a property of the input question, meaning it remains almost identical regardless of which answer the model generates. Adding this score shifts the certainty of the entire question up or down but leaves the internal ranking of candidate answers untouched. This makes input level detectors function like standard math adjustments: they might flag a difficult question, but they cannot help decide which specific answer to trust. A useful scoring method must look directly at the answer.

\textbf{The Latent Correctness Readout:}
\label{sec:readout}
Let $\bm{h}_i^{\ell}(\tilde{\bm{q}})$ denote the hidden state at layer $\ell$ for answer position $i\in\mathcal{A}$. We average these states over the answer and score correctness using a simple logistic tool
\begin{equation}
\label{eq:readout}
\bar{\bm{h}}^{\ell} = \frac{1}{|\mathcal{A}|} \sum_{i \in \mathcal{A}} \bm{h}_i^{\ell}(\tilde{\bm{q}}),
\qquad
r(\tilde{\bm{q}}) = \sigma\!\left(\bm{u}^\top \bar{\bm{h}}^{\ell} + b\right),
\end{equation}
where we fit the parameters $(\bm{u},b)$ on a small set of labeled examples. We call this $r$ the \textbf{Latent Correctness Readout} (LCR) and use it to rank answers and decide when to skip questions. The specific layer and mathematical settings are chosen using standard testing on the training data alone, consistently favoring the middle to late layers.

Two properties make this approach practical and revealing. First, it requires no updates to the main language model, fitting in seconds on a standard processor using cached data. Second, because it looks at the full internal state instead of the final probabilities, it varies across different possible answers and can reorder them. This is exactly what standard math adjustments and input level detectors fail to do. Because it adds no text generation and changes no model weights, any ranking ability it finds was already present in the frozen model.

\textbf{Why the readout works:}
\label{sec:why}
Previous research suggests that diffusion certainty on clean text measures how consistent the generated answer is with itself rather than whether it is actually correct \citep{li2026confidence}. This idea explains the failure we observe. When given text with a typo, the model fixes the prompt into a coherent problem and solves it consistently. Therefore, certainty remains high even though the answer is wrong compared to the user's actual intent. The final output certainty simply measures this self consistency. In contrast, the internal hidden states still separate correct answers from incorrect ones, and our tool reads this deeper signal. This also explains why standard math adjustments fail, because noise destroys the actual ranking ability rather than just creating an excess of certainty.

\section{Experimental Setup}
\label{sec:exp}

\paragraph{Models, tasks and noise:}
We evaluate LLaDA-8B-Instruct and DREAM-7B-Instruct with their standard confidence-ordered samplers, together with LLaMA-3-8B~\citep{llama3} and Qwen-2.5-7B~\citep{qwen2p5} as matched autoregressive controls. Fine-tuned baselines update only LoRA adapters~\citep{lora2022} of rank $16$, while base weights and all probes remain frozen at inference. The core study uses GSM8K and ARC-Challenge, and the generality analysis extends it to nine further benchmarks. Prompts are corrupted by character transposition, keyboard substitution, character insertion and word deletion at severities $\eta\in\{0.05,0.10,0.15,0.30\}$, and every corrupted prompt has a clean counterpart, which allows matched-pair testing. All experiments ran on two RTX 4090 GPUs.

\paragraph{Metrics:}
The primary metric is the selective AUROC between confidence and correctness, which is independent of accuracy. We also report AURC, a deployment composite that additionally depends on accuracy, and ECE over $15$ bins on the noisy set. Exact-match accuracy is a constraint that every intervention must preserve. Confidence is the geometric mean of token confidences given in Eq.~\ref{eq:conf}.

\paragraph{Protocol and splits:}
The two axes use disjoint data. On the accuracy axis, LoRA methods are trained on $200$ GSM8K prompts under transposition noise at $\eta=0.15$ and scored on a separate $100$-prompt cell. On the reliability axis, every confidence method, namely TS, DATS, the corruption probe and LCR, is fitted on a labeled calibration split of $200$ prompts and scored once on a disjoint held-out test split of $400$ prompts, with LCR hyperparameters chosen by five-fold cross-validation inside the calibration split. No test label is ever seen during fitting, and in deployment the operating threshold $\tau$ is likewise fixed on the calibration split to reach a target coverage. Throughout, $P$ denotes the empirical probability that a bootstrap effect exceeds zero over $10^4$ resamples of the test examples at a fixed noise draw. Variation across independent noise draws is reported separately below.

\paragraph{Baselines:}
We compare LCR against the base model with its raw geometric-mean confidence, the consistency baseline, post-hoc calibration by TS and DATS, the corruption probe used directly as a detector, ten same-budget scalar signals such as token entropy and answer length together with their combinations, and two sampling-based estimators, self-consistency agreement and semantic entropy.

\section{Results}
\label{sec:results}

\subsection{Accuracy is largely recovered by a simple baseline}
\label{sec:accuracy}
Basic matching training recovers most of the accuracy lost to noise. It raises accuracy from 0.370 to 0.490 on the accuracy testing set and from 0.372 to 0.405 on the larger reliability split. Three special diffusion training methods we designed did not reliably beat this basic approach. Because a standard training objective largely solves the accuracy problem, the rest of our evaluation focuses specifically on certainty reliability.

\subsection{Reliability degrades under noise and a readout recovers it}
\label{sec:reliability}
Table~\ref{tab:calib} reports the reliability results. The base model has poor calibration and barely separates correct from incorrect answers. The standard mathematical adjustments behave exactly as our structural argument predicts. They reduce the calibration error to 0.133 but leave the ranking order entirely unchanged. An input level error detector is the most perfectly calibrated score in the table, yet it ranks answers near random chance while easily separating clean from noisy text. A score can therefore be perfectly calibrated and still remain completely useless for deciding which answers to trust, because it flags noisy inputs rather than wrong answers.

\begin{figure}[t]
\centering
\includegraphics[width=\columnwidth]{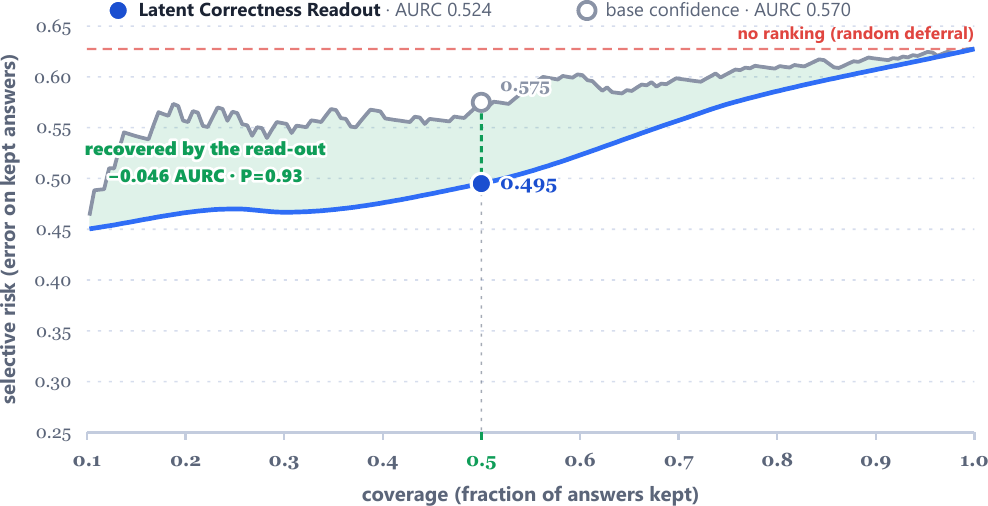}
\caption{Selective prediction under noise on GSM8K. Our tool cuts the error on kept answers significantly.}
\label{fig:riskcoverage}
\end{figure}

Our tool raises ranking quality to 0.665 on the same training data without touching a single model weight. We use this 0.665 held out estimate as our primary performance number. This number establishes that the ranking order destroyed at the output is still recoverable from the internal representation. Standard math adjustments could never show this. At 50 percent coverage the error on kept answers falls from 0.575 to 0.495 (Figure~\ref{fig:riskcoverage}). The basic training method reaches a lower overall risk, but it does so through accuracy rather than ranking. Stacking both gives the best single pass system available in our tests.

\textbf{Sampling ranks better and that is not a defeat:}
Generating extra answers ranks correctness better than any single pass score. Ten extra generations reach 0.842, and a single extra generation reaches 0.711, which sits above our tool's 0.665. We are not proposing our tool as the primary method a practitioner with spare compute should choose. We state the ordering plainly because from two extra generations upward, sampling alone dominates every variant we tested.

\begin{figure}[t]
\centering
\includegraphics[width=\columnwidth]{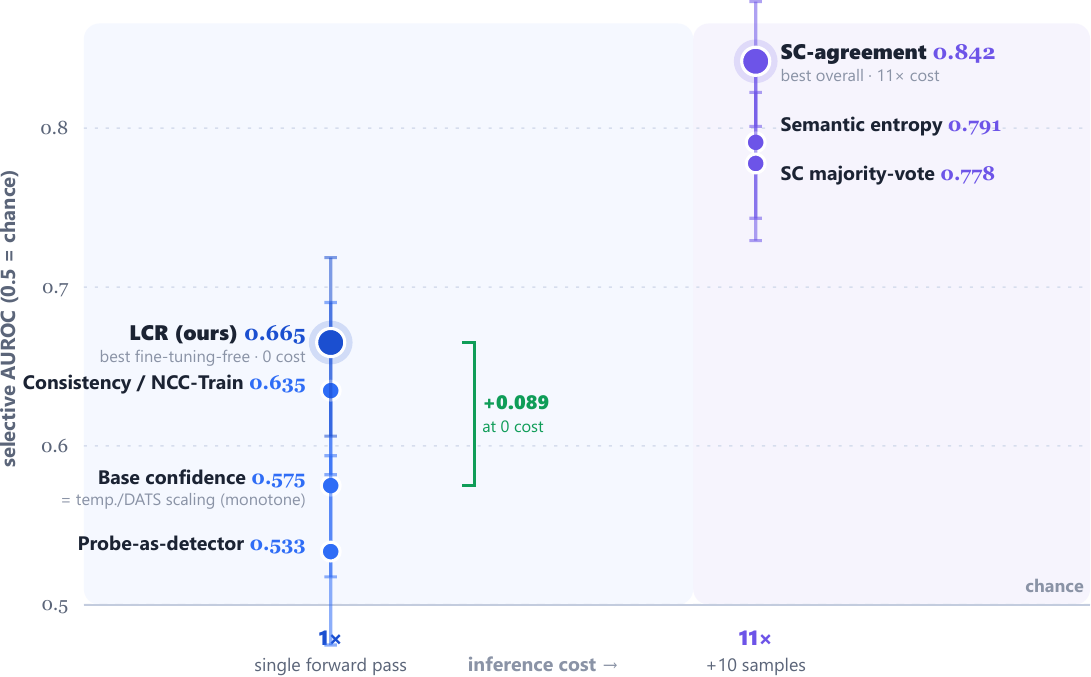}
\caption{Inference cost against ranking ability on GSM8K under noise. Single pass methods sit on the left edge and ten sample estimators on the right. Intermediate budgets appear in Table~\ref{tab:calib}.}
\label{fig:cost}
\end{figure}

Our tool earns its place here as evidence rather than as a final product. It is fitted on frozen states, updates nothing, and adds zero generation steps. Therefore, the ranking it recovers can only come from information the model already held and failed to report. A sampling method cannot establish this because it creates new information rather than revealing hidden information. Two practical notes follow from this. Our tool is the only option listed that adds no generation, making it useful when an answer must be produced in one pass. At small budgets the two methods compose well. Averaging our tool with one sampled answer reaches 0.754 against 0.711 for sampling alone. At two generations the pair still leads slightly, but from three upward sampling alone is better. Every extra generation requires a full model run, making compute efficiency highly relevant (Figure~\ref{fig:cost}).

\begin{table}[t]
\centering
\small
\caption{\small Confidence reliability under noise on GSM8K. Bold marks the best value overall. Italics mark the best at a given budget.}
\setlength{\tabcolsep}{3pt}
\resizebox{\columnwidth}{!}{%
\begin{tabular}{lccccc}
\toprule
Method & Gen & ECE$\downarrow$ & AUROC$\uparrow$ & AURC$\downarrow$ & Acc \\
\midrule
Base model & 1 & 0.610 & 0.575 & 0.570 & 0.372 \\
\midrule
\multicolumn{6}{l}{\emph{Rescale confidence}}\\
\;+ TS & 1 & 0.133 & 0.575 & 0.570 & 0.372 \\
\;+ DATS & 1 & 0.133 & 0.575 & 0.570 & 0.372 \\
\midrule
\multicolumn{6}{l}{\emph{Recover accuracy or route corruption}}\\
Consistency & 1 & 0.581 & 0.637 & 0.478 & 0.405 \\
\;+ NCC Train & 1 & 0.554 & 0.632 & 0.471 & \best{$\mathbf{0.430}$} \\
Probe as detector & 1 & \best{$\mathbf{0.012}$} & 0.533 & 0.586 & 0.372 \\
\midrule
\multicolumn{6}{l}{\emph{Sample the model}}\\
Self consistency ($k{=}1$) & 2 & {0.394} & 0.711 & {0.492} & 0.372 \\
Self consistency ($k{=}2$) & 3 & {0.286} & 0.771 & {0.445} & 0.372 \\
Semantic entropy ($k{=}10$) & 11 & 0.373 & 0.791 & 0.425 & 0.372 \\
Self consistency ($k{=}10$) & 11 & {$\mathit{0.117}$} & \best{$\mathbf{0.842}$} & \best{$\mathbf{0.383}$} & 0.372 \\
\;SC majority vote & 11 & 0.183 & 0.778 & 0.386 & {$\mathit{0.415}$} \\
\midrule
\multicolumn{6}{l}{\emph{Read out correctness}}\\
\textbf{LCR} & 1 & 0.123 & 0.665 & 0.524 & 0.372 \\
\textbf{Consist + LCR} & 1 & 0.120 & {$\mathit{0.688}$} & {$\mathit{0.442}$} & 0.405 \\
\textbf{LCR + SC} ($k{=}1$) & 2 & {{$\mathit{0.181}$}} & {$\mathit{0.754}$} & {{$\mathit{0.459}$}} & 0.372 \\
\midrule
\multicolumn{6}{l}{\emph{Autoregressive control LLaMA 3 8B}}\\
Base confidence & 1 & 0.576 & 0.628 & 0.606 & 0.315 \\
\;+ LCR & 1 & 0.091 & 0.721 & 0.539 & 0.315 \\
\bottomrule
\end{tabular}}
\label{tab:calib}
\end{table}

\textbf{Results across independent noise draws:}
The initial statistical values resample test examples at a fixed noise draw, which understates the variability of the noise itself. We therefore repeated the entire pipeline over four independent noise seeds. On ARC the readout gains 0.078 on average and remains positive on all four seeds. On GSM8K it gains 0.073 and is positive on three. The single negative seed traces to model selection during training rather than harsh corruption. The effect is reproducible on ARC and variable on GSM8K, and it relies on a selection step a deployer cannot easily audit without labels. This variability is one reason we pair the tool with the diagnostic check below.

\subsection{The gain needs the hidden state}
\label{sec:hidden}
Ten cheaper single pass signals fitted on the same data stay below a ranking score of 0.590. Corruption and correctness are distinct variables, and no single number carries the correctness signal. The correctness signal also lives in the generated answer rather than the prompt. A tool restricted to prompt hidden states reaches only 0.647, whereas adding the answer states raises it to 0.720.

\subsection{Diffusion confidence loses more of its ranking}
\label{sec:ar}
Under identical noise both standard controls remain usable, while both diffusion models fall to the edge of random chance. All four cross family pairings favor the standard model under noise. On clean text the comparison is not consistent, as one standard model ranks below LLaDA. What we can reliably demonstrate is the specific deficit that noise induces.

The limits of this comparison should be stated plainly. The statistical tests resample examples with the four models held fixed, so this supports a claim about these specific systems rather than a broad claim about all architectures. The certainty definitions also differ by construction. Diffusion scores aggregate maximum values recorded during generation, while standard scores aggregate standard probabilities. We report this comparison as a control, and our main internal arguments do not depend on it.

\textbf{Concentration is a symptom of the scale:}
The visible form of the failure is that diffusion certainty is clustered against its ceiling. Under noise LLaDA reports high certainty on 99.2 percent of its errors where LLaMA 3 8B reports it on 2.2 percent. It is tempting to read this concentration as the substance of the problem, but it is not.

The fraction of errors above a fixed cutoff changes easily with standard math adjustments, whereas a true ranking decision does not. Applying a basic temperature adjustment removes this clustering entirely, driving the LLaDA error fraction from 0.992 down to 0.004 while leaving the ranking score completely unchanged. A statistic that changes by two orders of magnitude without altering a single model decision cannot be what breaks the ability to skip questions. Asking the same question at a matched operating point rather than a fixed cutoff confirms this. Most of the apparent gap between the model families is simply the location of the cutoff rather than the shape of the distribution. The true issue is the ranking deficit.

\textbf{The tool is a general instrument:}
The same tool fitted on standard hidden states lifts LLaMA 3 8B from 0.628 to 0.721 under noise. A linear correctness direction therefore survives noise in both families. The tool is a general instrument rather than a diffusion specific repair. What separates the families is not whether this internal truth exists, since both carry it, but how much of it reaches the number the model actually reports. Under noise the standard models retain enough signal for a cutoff to act on, while the diffusion models retain very little.

\textbf{Does the decoding schedule manufacture the effect:}
A diffusion sampler fills whichever hidden position is currently most certain, so the scores we record are maximums by design. It is fair to ask whether this design alone produces what we report. For the ranking claim, the design already answers this. The generation process is identical on clean and noisy prompts, so a mechanism constant across both conditions cannot produce a difference between them.

Further observations limit this effect. The sampler discards nothing during generation, meaning it fixes the order in which positions are measured rather than which measurements survive. Standard models also take a maximum over their vocabulary. Most directly, scoring an answer by its weakest recorded word rather than its average still results in a poor ranking score of 0.573. No mathematical average we tried recovers the lost signal.

\begin{table}[t]
\centering
\small
\caption{\small Where the families differ on the held out test split. The large fixed cutoff gap is a surface artifact completely removed by standard temperature scaling.}
\setlength{\tabcolsep}{3.5pt}
\resizebox{\columnwidth}{!}{%
\begin{tabular}{llcccccc}
\toprule
& & \multicolumn{2}{c}{err$>$0.95} & \multicolumn{2}{c}{err$>\tau_{80}$} & \multicolumn{2}{c}{AUROC} \\
\cmidrule(lr){3-4}\cmidrule(lr){5-6}\cmidrule(lr){7-8}
Model & Fam & Cln & Noi & Cln & Noi & Cln & Noi \\
\midrule
LLaDA 8B & D & 1.000 & 0.992 & 0.739 & 0.777 & 0.664 & 0.575 \\
DREAM 7B & D & 0.778 & 0.180 & 0.722 & 0.775 & 0.565 & 0.540 \\
LLaMA 3 8B & A & 0.024 & 0.022 & 0.549 & 0.748 & 0.741 & 0.628 \\
Qwen 2.5 7B & A & 0.084 & 0.089 & 0.723 & 0.756 & 0.624 & 0.612 \\
\midrule
\multicolumn{2}{l}{D minus A pooled} & +.835 & +.531 & +.095 & +.024 & -.068 & -.063 \\
\midrule
\multicolumn{2}{l}{\emph{invariant}} & \multicolumn{2}{c}{no} & \multicolumn{2}{c}{yes} & \multicolumn{2}{c}{yes} \\
\bottomrule
\end{tabular}}
\label{tab:confmass}
\end{table}

\begin{figure*}[t]
\centering
\includegraphics[trim=26 0 0 66,clip,width=\textwidth]{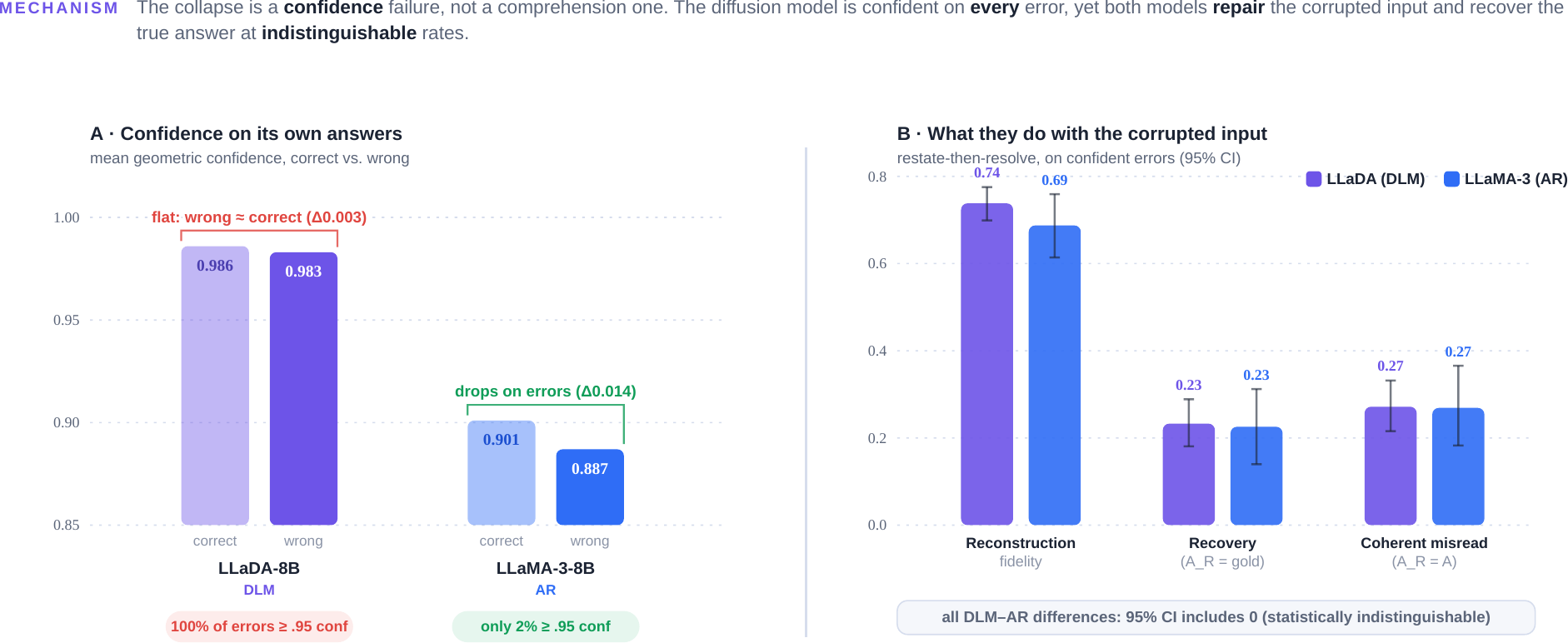}
\caption{Confidence fails where comprehension does not. The diffusion model scores wrong and right answers alike on the raw scale.}
\label{fig:mechanism}
\end{figure*}

\subsection{Generalization and limitations}
\begin{figure}[t]
\centering
\includegraphics[width=\columnwidth]{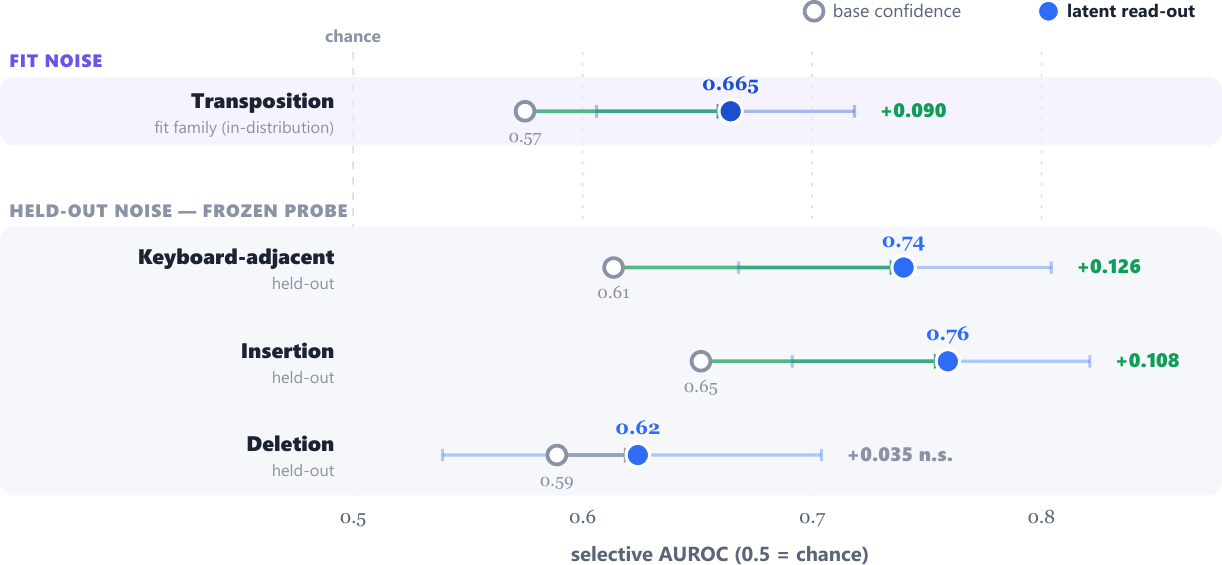}
\caption{Transfer to unseen noise types on GSM8K.}
\label{fig:transfer}
\end{figure}

\textbf{Unseen noise and tasks:}
Fitted once on clean and swapped text, the tool transfers to keyboard substitution and character insertion (Figure~\ref{fig:transfer}). Missing words shorten the prompt and disrupt word alignment, remaining a boundary case. The tool remains safe under unanticipated noise within a supported task. On a new task like ARC Challenge it restores 0.105 ranking quality, but only after a short fit on the new data. A deployed system must reuse one tool across noise conditions but refit it for each new task.

\textbf{Where the tool does not transfer:}
\label{sec:predict}
Across eleven benchmarks the tool helps on five and gives no test gain on the other six. On WinoGrande it actively hurts performance by 0.070. This is the tool's clearest limitation and we report it directly. Task category does not fully explain this split. On CommonsenseQA a strong training fit collapsed entirely on the testing set, showing the tool can latch onto dataset specific structures instead of a stable correctness direction. Finding a stable correctness direction under noise is a property of some tasks and not a general guarantee.

These failures are at least visible before deployment. Using the training data alone we calculate a margin score comparing our tool to the base model. All five successful tasks score high on this margin and all six failed tasks score low (Table~\ref{tab:predict}). We claim this only as a screening heuristic, and it is a weak one. The separating window is narrow, eleven tasks is a small sample, and the scores only correlate moderately with the realized gain. Its main practical use is asymmetric. WinoGrande is the only task the tool would actively hurt, and the rule successfully screens it out.

\begin{figure}[t]
\centering
\includegraphics[width=\columnwidth]{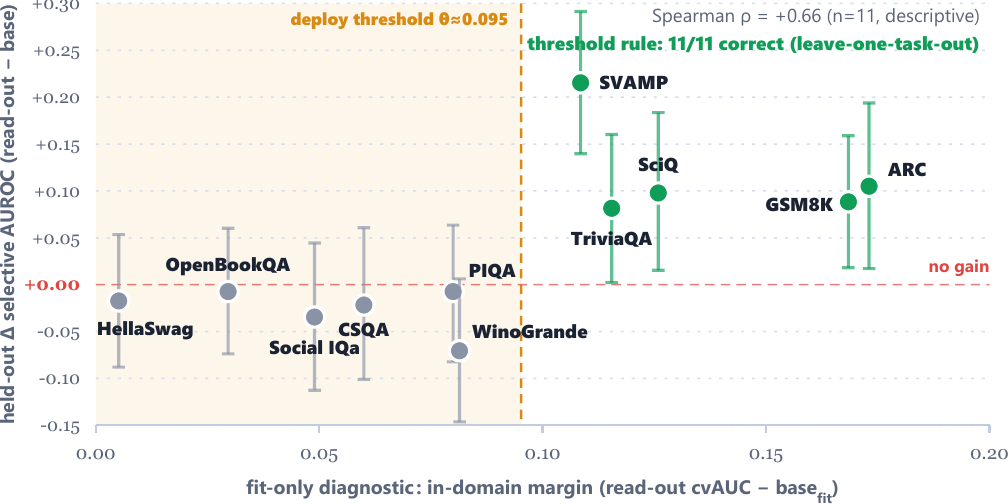}
\caption{Screening before deployment. One threshold separates the tasks the tool helps from those it does not.}
\label{fig:generality}
\end{figure}


\begin{table}[t]
\centering
\small
\caption{\small Screening tasks with training data alone. A threshold separates the successful tasks from the null ones.}
\setlength{\tabcolsep}{4pt}
\resizebox{\columnwidth}{!}{
\begin{tabular}{lcccc}
\toprule
Task & Base & LCR cv & $m$ & Test $\Delta$ \\
\midrule
ARC Challenge \citep{clark2018think} & 0.451 & 0.624 & {$\mathbf{+0.173}$} & +0.105 \\
GSM8K \citep{cobbe2021training} & 0.619 & 0.788 & {$\mathbf{+0.168}$} & +0.090 \\
SciQ \citep{welbl2017crowdsourcing} & 0.540 & 0.666 & {$\mathbf{+0.126}$} & +0.098 \\
TriviaQA \citep{joshi2017triviaqa} & 0.685 & 0.801 & {$\mathbf{+0.115}$} & +0.082 \\
SVAMP \citep{patel2021are} & 0.549 & 0.658 & {$\mathbf{+0.109}$} & +0.215 \\
\midrule
WinoGrande \citep{sakaguchi2020winogrande} & 0.514 & 0.595 & +0.081 & -0.070 \\
PIQA \citep{bisk2020piqa} & 0.584 & 0.664 & +0.080 & -0.007 \\
CommonsenseQA \citep{talmor2019commonsenseqa} & 0.556 & 0.616 & +0.060 & -0.022 \\
Social IQa \citep{sap2019social} & 0.509 & 0.558 & +0.049 & -0.034 \\
OpenBookQA \citep{mihaylov2018can} & 0.612 & 0.641 & +0.030 & -0.007 \\
HellaSwag \citep{zellers2019hellaswag} & 0.607 & 0.612 & +0.005 & -0.017 \\
\bottomrule
\end{tabular}}
\label{tab:predict}
\end{table}

\textbf{Further ablations:}
Analyses in the appendix confirm that negative routing results hold at larger sample sizes, that the tool works with as few as 25 training examples, and that the improvements are robust to binning artifacts.
\section{Related Work}
\label{sec:related}

\textbf{Diffusion language models:}
Masked diffusion models \citep{d3pm2021,sedd2024,mdlm2024} generate text through repeated noise removal. Recent work scales these methods to larger models like LLaDA \citep{llada2025} and DREAM \citep{dream2025} alongside other improvements \citep{cdlm2025,khadangi2025}. This research focuses on text quality and generation speed rather than how the models handle input errors.

\textbf{Robustness to text noise:}
Previous methods to handle text noise include smoothing goals \citep{vat2018,smart2020,freelb2020,rdrop2021,uda2020,meanteacher2017}, internal state defenses \citep{dne2021,infobert2021}, typo corrections \citep{pruthi2019,roben2020,textfooler2020}, and adversarial training \citep{lat2024}. These were built for standard models to improve accuracy. Our basic training method comes from this group. We focus on the certainty problem these methods ignore, because fixing accuracy is the easier challenge for diffusion models.

\textbf{Confidence in diffusion language models:}
Existing studies on diffusion certainty only look at clean text. \citet{li2026confidence} note a certainty paradox on math problems where the raw score is poorly calibrated but still ranks answers well. This means standard math adjustments work well on clean text. We successfully reproduce this exact finding. Their adjustment lowers our calibration error from 0.410 to 0.015 on clean text while keeping the ranking score steady at 0.663. However, under noisy conditions this same adjustment leaves the ranking frozen at 0.571. The noisy environment causes the failure, not the adjustment method. Our overall clean ranking is slightly lower than their math results, likely due to a smaller generation budget and different scoring math. We build on their idea that diffusion certainty measures internal consistency rather than input accuracy. 

\textbf{Calibration and the representation confidence gap:}
We use standard methods to measure calibration and question skipping \citep{naeini2015,guo2017,geifman2017,hendrycks2017baseline}. For standard models, researchers already know that internal states predict correctness better than final output scores \citep{burns2022discovering,azaria2023internal,chen2024inside,kossen2024semantic,xiong2024can,kadavath2022,jiang2021}. They also know that generating extra answers improves ranking \citep{wang2023selfconsistency,kuhn2023semantic}. This extra generation is very expensive for diffusion models because each answer requires many steps. We do not claim to have discovered the gap between internal truth and final certainty. Our own standard models reproduce it. We add a careful measurement of this gap under text errors where it becomes most severe. Very high certainty scores might seem to mean the model has no low certainty errors to skip. We prove this surface reading is false. A simple math adjustment removes the high scores entirely without changing any actual model decisions. What survives this adjustment is a deep loss of ranking ability, an issue the field understands but has not yet measured in noisy environments.

\vspace{-10pt}
\section{Conclusion}
\label{sec:conclusion}

Under real world text errors diffusion models show a split behavior. Internal states detect errors highly accurately while final certainty stays near maximum and fails to separate right from wrong answers. Since basic training easily recovers lost accuracy, certainty reliability becomes the main roadblock for safe deployment. Concentrated high certainty is the most obvious but least important symptom. A simple math adjustment removes it entirely without altering any model decisions. Once adjusted the massive gap between diffusion and standard models shrinks significantly, revealing a real loss of ranking order. This loss favors standard models in all our noisy comparisons. Basic math adjustments and error detectors cannot fix this because they do not reorder answers.
Our simple extraction tool partially fixes this to prove the internal signal exists. It updates no model weights and adds zero generation steps, meaning any recovered ranking comes from information the model already possessed but failed to report. Its limits are clear. It requires task specific training data, fails to improve six out of eleven benchmarks, and falls behind methods that generate two or more extra answers. The information needed to know when a diffusion model is wrong already exists inside it. What is missing is the proper connection between that internal truth and the certainty score the model actually reports.

{\small
\bibliography{ref}
}
\clearpage
\appendix
\renewcommand{\thesection}{SARabic{section}}

\noindent This appendix collects the full derivations of the routing attempts,
the extended autoregressive-vs-diffusion analysis, the localization and frozen-transfer
controls, and the ablations, referred to from the main paper. Section, table, figure, and
equation numbers here are prefixed with ``S''; references without the ``S'' prefix (e.g.\ ``Table~2'') point back into the main paper.

\section{Routing attempts: full derivations}

\subsection{Routing attempt I---probe-gated confidence penalty (Ncc-Train)}
\label{sec:ncc-train}
The training component makes the model \emph{express} the uncertainty it already
perceives. During fine-tuning we have gold answers, so we can be precise: we penalize
committed confidence only where (i) the probe flags corruption and (ii) the model is
actually getting the token wrong. Let $\kappa_i=\max_v f_\theta(v\mid\tilde{\bm{q}})_i$ be the
peak (commit) probability and $p_i^{\mathrm{gold}}=f_\theta(y_i\mid\tilde{\bm{q}})_i$ the
probability mass on the gold token. Define a per-position \emph{hedge weight}
\begin{equation}
\label{eq:hedge}
m_i \;=\; g_i \cdot \big(1-p_i^{\mathrm{gold}}\big),
\end{equation}
which is large exactly when a position is both corruption-influenced ($g_i$ high) and
likely wrong ($p_i^{\mathrm{gold}}$ low), and small on confident-correct positions
($p_i^{\mathrm{gold}}$ high) so it does not punish justified confidence. The
calibration penalty pushes down peak confidence in proportion to the hedge weight:
\begin{equation}
\label{eq:cal}
\mathcal{L}_{\mathrm{cal}}(\theta)=
\E_{i\in\mathcal{A}}\big[\,m_i\,\cdot\,\kappa_i\,\big].
\end{equation}
Intuitively, on corruption-influenced positions where the model is about to be
confidently wrong, Eq.~\ref{eq:cal} spreads probability mass away from the (wrong) peak,
raising entropy \emph{selectively}. This is a corruption-gated, correctness-aware
variant of the confident-output regularizers used for classifier
calibration~\citep{pereyra2017,mukhoti2020}, but tied to the DLM's own internal
corruption detector rather than applied uniformly. The full training loss is
\begin{equation}
\label{eq:full}
\mathcal{L}=
\underbrace{\mathcal{L}_{\mathrm{sup}}(\tilde{\bm{q}})}_{\text{answer}}
+\;\lambda_{\mathrm{cons}}\,\mathcal{L}_{\mathrm{cons}}
+\;\lambda_{\mathrm{cal}}\,\mathcal{L}_{\mathrm{cal}},
\end{equation}
where $\mathcal{L}_{\mathrm{sup}}$ is the weighted masked negative log-likelihood on the noisy
input and $\lambda_{\mathrm{cons}},\lambda_{\mathrm{cal}}$ balance accuracy recovery and
calibration. Only LoRA adapters are updated.

\subsection{Routing attempt II---risk-deferred commitment (Ncc-Decode)}
\label{sec:ncc-decode}
The decoding component is training-free and exploits the commitment schedule directly.
The standard rule commits, at each reverse step, the positions with the highest raw
confidence $\kappa_i$. We instead commit by a \emph{risk-discounted} priority
\begin{equation}
\label{eq:rdc}
\tilde{\kappa}_i=\kappa_i\cdot\big(1-\gamma\,g_i\big),\qquad \gamma\in[0,1],
\end{equation}
so that a position that is confidently predicted \emph{but} flagged as
corruption-influenced is committed \emph{later}. Deferring risky positions lets the
bidirectional decoder first lock in the clean, unambiguous parts of the answer; the
deferred positions are then resolved with strictly more committed context, which both
improves their accuracy and makes their eventual commit-confidence more meaningful. Two
properties matter. First, Eq.~\ref{eq:rdc} changes the \emph{order} of commitment, not
the vocabulary distribution, so it is a pure scheduling intervention. Second, it is
\emph{impossible in an AR model}, which has a single fixed decoding order; it is a
control lever unique to diffusion. NCC-Decode composes with any DLM and with
NCC-Train.

\begin{algorithm}[t]
\caption{NCC decoding with risk-deferred commitment}
\label{alg:decode}
\begin{algorithmic}[1]
\REQUIRE noisy prompt $\tilde{\bm{q}}$; DLM $f_\theta$; probe $(\bm{w},b,\ell^\star)$;
schedule $\{k_r\}$; discount $\gamma$
\STATE initialize canvas $\bm{x}\leftarrow[\texttt{MASK}]^{L}$
\FOR{$r=1$ \TO $R$}
  \STATE $p_i \leftarrow f_\theta(\cdot\mid \bm{x},\tilde{\bm{q}})_i$ for masked $i$;\;
         $\kappa_i\leftarrow\max_v p_i(v)$
  \STATE $g_i\leftarrow\sigma\!\big(s(\bm{w}^\top \bm{h}_i^{\ell^\star}-b)\big)$
         \hfill// corruption score
  \STATE $\tilde{\kappa}_i\leftarrow\kappa_i(1-\gamma g_i)$
         \hfill// risk-deferred priority (Eq.~\ref{eq:rdc})
  \STATE commit the $k_r$ masked positions with largest $\tilde{\kappa}_i$;\;
         record $\kappa_i^\star=\kappa_i$ at commit
\ENDFOR
\STATE \textbf{return} filled $\bm{x}$ and confidence $c=\big(\prod_{i\in\mathcal{A}}\kappa_i^\star\big)^{1/|\mathcal{A}|}$
\end{algorithmic}
\end{algorithm}

\section{Extended results and analyses}
\paragraph{Fit and notation conventions.} Unless stated otherwise, the extended
analyses in this section fit the readout on the combined clean$+$noisy calibration
pool---the labeled selective AUROC of $0.720$ reported alongside the main results---whereas
the main-text headline uses the deployment-faithful noisy-only fit ($0.665$; the
main-paper calibration table). The two protocols agree on every qualitative
conclusion and differ only in the size of the reported gain, as discussed in the main text.
Where a paired-bootstrap $P{=}1.000$ appears below, it means the exceedance probability rounds
to $1.000$ at three decimals---an empirical bound from a finite resample, not a claim of
certainty.

\subsection{The signal is localized to the answer, not the input}
The same-budget table rules out cheaper \emph{scalars}; a further question is \emph{where} in
the representation the readout's signal lives. Does it read answer-level correctness, or merely
how hard---or how corrupted---the input was? We separate the two by pooling hidden states over
the \emph{prompt} tokens instead of the answer tokens, under three progressively stricter
controls, all fit with the identical protocol and scored on the same $n{=}400$ noisy test set
(Table~\ref{tab:localize}). A probe on the prompt representation from the \emph{same} forward
pass the readout uses (the committed prompt-plus-answer sequence) reaches $0.676$. But DLM\
attention is bidirectional, so those prompt tokens have already attended to the answer---this is
not a clean input-only probe. We therefore re-run the model over the \emph{prompt alone}, with
no answer appended; the prompt tokenization is byte-identical, so the only thing removed is the
answer's bidirectional influence. This truly input-only representation reaches $0.647$ (CI
$[0.591,0.700]$), numerically above the raw-confidence scalar ($0.575$)---the model's internal
representation of a corrupted question already encodes some of its difficulty---but well below
the readout.

The key quantity is the increment the answer adds. The answer representation ($0.720$)
exceeds the truly input-only representation by $+0.073$ (paired bootstrap $P{=}1.000$, CI
$[+0.032,+0.115]$): a signal that appears only once the answer is written and is absent from
\emph{any} representation of the input, hence specifically answer-level rather than a restatement of
input difficulty. Letting the prompt tokens attend to the answer (the bidirectional pass)
recovers only $+0.029$ of this gap ($P{=}0.99$), leaving the answer representation still
significantly ahead ($+0.044$, $P{=}0.99$). Input difficulty thus explains part of the readout's
gain over confidence, but a significant, irreducible part requires reading the answer itself---the
readout's edge is localized to the answer tokens.

\begin{table}[t]
\centering
\small
\setlength{\tabcolsep}{4pt}
\resizebox{\columnwidth}{!}{%
\begin{tabular}{lcc}
\toprule
Representation (same budget) & Sel.\ AUROC & $95\%$ CI \\
\midrule
Raw confidence (scalar)                       & $0.575$ & $[0.517,0.633]$ \\
Input-only prompt (no answer, $4096$-d)       & $0.647$ & $[0.591,0.700]$ \\
Joint-pass prompt (attends answer, $4096$-d)  & \sbest{$\mathit{0.676}$} & $[0.621,0.728]$ \\
\midrule
\textbf{Answer readout (LCR, $4096$-d)}     & \best{$\mathbf{0.720}$} & $[0.668,0.771]$ \\
\bottomrule
\end{tabular}}
\caption{Localizing the readout's signal (GSM8K, transposition $\eta{=}0.15$, $n_{\mathrm{test}}{=}400$, identical fit protocol). Pooling \emph{prompt} tokens from the model's forward pass gives $0.676$, but bidirectional attention lets them absorb the answer. Re-running over the \emph{prompt alone} isolates a genuinely input-only representation at $0.647$, above the confidence scalar; the answer readout still adds $+0.073$ (paired $P{=}1.000$, CI $[+0.032,+0.115]$). Bold marks the best representation, italics the strongest alternative.}
\label{tab:localize}
\end{table}

\subsection{The confidence degradation is disproportionately a diffusion phenomenon}
\begin{figure}[t]
\centering
\includegraphics[trim=95 0 10 108,clip,width=\columnwidth]{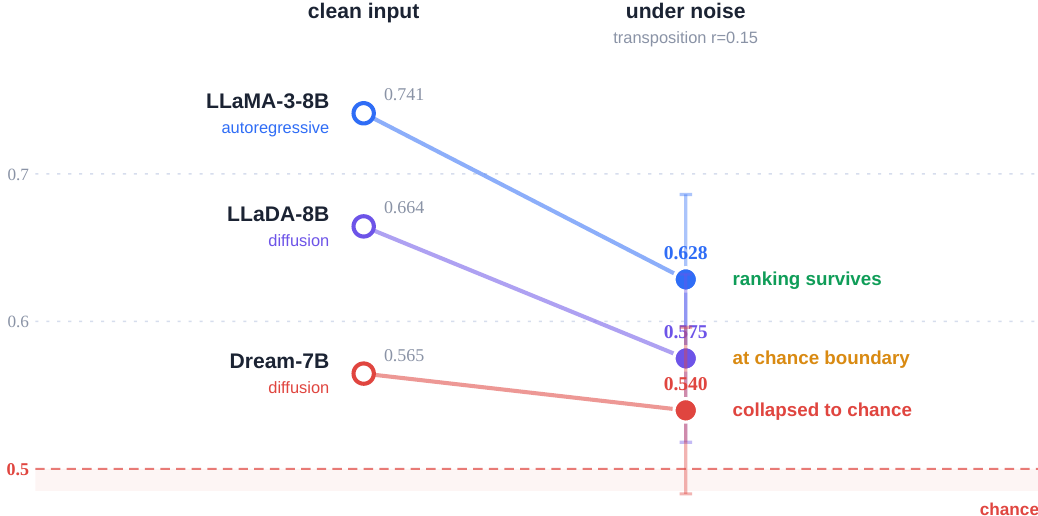}
\caption{Base confidence as a correctness ranker, clean against noisy. LLaMA-3 keeps a signal that clears chance under noise, whereas both DLM{}s fall to the edge of usefulness and Dream's interval covers chance. Qwen behaves like LLaMA-3\ and is listed in Table~\ref{tab:ar}.}
\label{fig:arcollapse}
\end{figure}

Is this confidence degradation a property of DLM{}s, or merely of hard inputs that would break any
model's confidence? We rerun the identical protocol---same prompts, transposition noise
($\eta{=}0.15$), geometric-mean confidence, held-out selective AUROC ($n_{\mathrm{test}}{=}400$)---on
an autoregressive model of the same scale, LLaMA-3-8B~\citep{llama3}, and---to guard against
reading too much into a single AR---a second, architecturally distinct one, Qwen-7B~\citep{qwen2p5};
we compare their base-confidence discrimination to LLaDA's and Dream's
(Table~\ref{tab:ar}; Figure~\ref{fig:arcollapse} visualizes the first AR and the two DLM{}s). Under
noise both AR models' confidence remains a better-than-chance ranker of correctness (LLaMA-3
selective AUROC $0.628$, CI $[0.570,0.686]$; Qwen $0.612$, CI $[0.554,0.668]$; both CIs exclude
$0.5$), whereas the DLM{}s sit at the chance boundary
($0.575$, CI $[0.518,0.632]$; $0.540$, CI $[0.483,0.596]$, the latter including $0.5$). A paired
bootstrap over the shared prompts sharpens what is and is not established: LLaMA-3's advantage is
significant over Dream ($+0.088$, CI $[+0.007,+0.171]$, $P{=}0.98$) but only directional over
LLaDA ($+0.053$, CI $[-0.024,+0.128]$, $P{=}0.91$, n.s.). We therefore make the careful claim
rather than the strong one---what is robust is that the AR{}s' confidence stays
above chance while both DLM{}s sit \emph{at} the chance boundary, not that either
AR is pairwise-separable from every individual DLM. The second AR, Qwen-7B, reproduces this
picture: its noisy base confidence also stays above the boundary ($0.612$, CI $[0.554,0.668]$) and is
statistically indistinguishable from LLaMA-3{} (paired difference $-0.016$, $P{=}0.33$), while its
own gaps over the DLM{}s are directional but do not reach significance (vs.\ Dream{} $+0.072$,
$P{=}0.96$; vs.\ LLaDA{} $+0.037$, $P{=}0.83$)---which we report rather than round up. The cross-family
statement thus does not rest on any single pairwise test. The AR{}s
are not immune---their confidence does degrade under noise---but they retain an above-chance signal where
the DLM{}s do not. This is consistent with the mechanism proposed in the main paper, which we probe
behaviorally
below (Table~\ref{tab:mech}): both families can reconstruct the corrupted problem when asked, but only
the DLM's confidence stays high on the resulting errors---consistent with confidence tracking the
internal consistency of the
solution it decodes rather than the fidelity of the input, whereas the AR's confidence still registers
the corruption. The readout helps wherever confidence is degraded---on LLaMA-3 it adds a significant
$+0.093$ under noise ($0.628{\to}0.721$, $P{=}0.996$) but nothing on clean inputs ($+0.016$, n.s.)---so
it is a general remedy; but the \emph{need} is greatest for diffusion models, whose confidence
degrades furthest.

\begin{table}[t]
\centering
\small
\begin{tabular}{llcc}
\toprule
Model & Type & Clean & Noisy \\
\midrule
LLaMA-3-8B & AR   & \best{$\mathbf{0.741}$} & \best{$\mathbf{0.628}$} \\
Qwen-7B       & AR   & $0.624$ & $0.612$ \\
LLaDA-8B      & DLM  & $0.666$ & $0.575$ \\
Dream-7B      & DLM  & $0.565$ & $0.540$ \\
\bottomrule
\end{tabular}
\caption{Base-model confidence as a correctness ranker (held-out selective AUROC, GSM8K, transposition $\eta{=}0.15$, $n_{\mathrm{test}}{=}400$). Under noise both autoregressive models retain ranking ability while both diffusion models fall to near chance, an edge that is significant against Dream\ and directional against LLaDA\ under a paired bootstrap. The restate-then-resolve probe (Table~\ref{tab:mech}) localizes the collapse to confidence rather than comprehension. Bold marks the strongest ranker in each column.}
\label{tab:ar}
\end{table}

\begin{figure*}[t]
\centering
\includegraphics[width=0.98\textwidth]{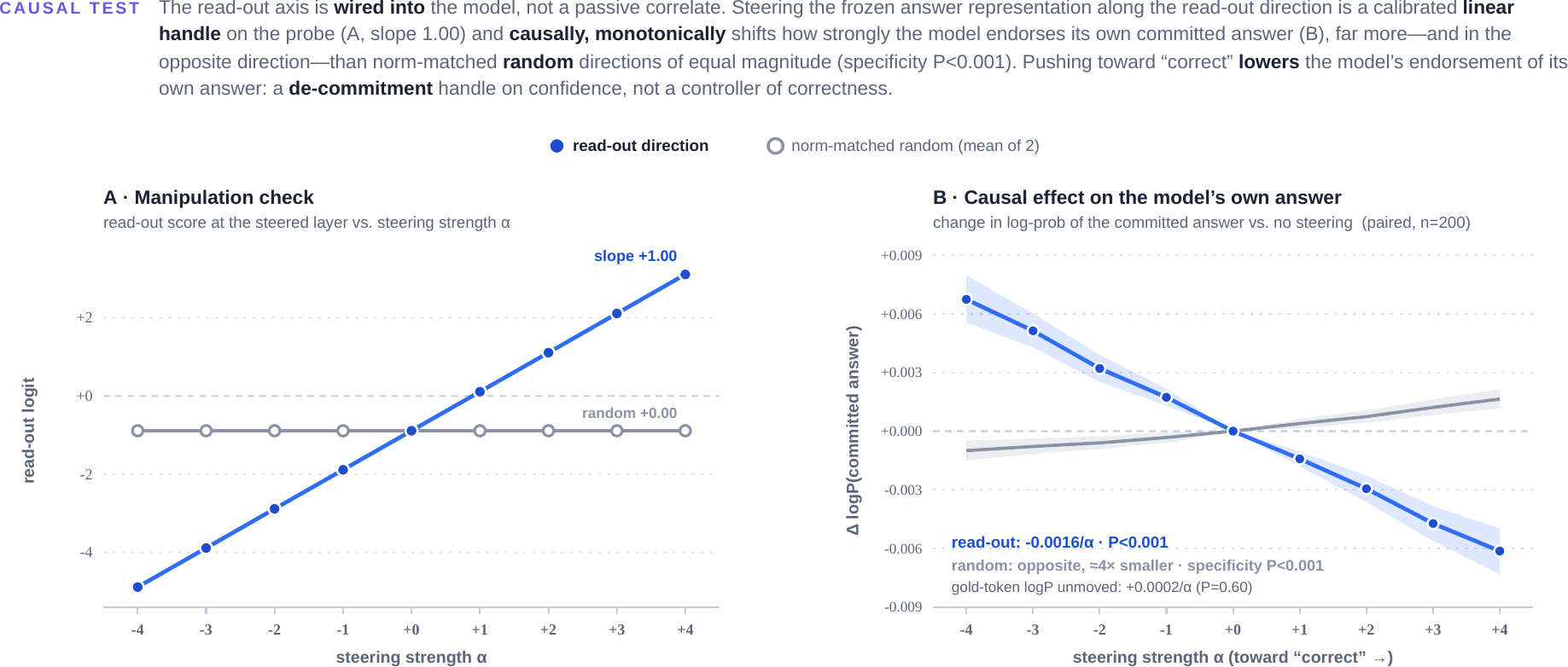}
\caption{\textbf{Causal intervention on the readout axis} (LLaDA-8B, GSM8K, transposition $\eta{=}0.15$, $n{=}200$ held-out noisy prompts). We add $\alpha\bm{d}$ along the readout's correctness direction to each answer-token residual at layer $\ell$ and sweep $\alpha$, measuring in a teacher-forced pass against norm-matched random controls. \textbf{(A)} The intervention is an exact linear handle on the readout (slope $1.00$), which fixes the scale of $\alpha$. \textbf{(B)} Steering toward the readout's ``correct'' side lowers the log-probability of the model's own committed answer ($-0.0016$ per $\alpha$, paired bootstrap $P{<}0.001$; bands are $95\%$ CIs), opposite in sign to the control, yet leaves the gold-token likelihood unmoved ($+0.0002$ per $\alpha$, $P{=}0.60$). Re-decoding under the same intervention does not move accuracy ($P{=}0.33$): the axis is a causal handle on \emph{confidence}, not on \emph{correctness}.}
\label{fig:causal}
\end{figure*}

\paragraph{A direct test: silent repair versus confident error.}
The main paper \emph{hypothesizes} a specific mechanism for the degradation---the model normalizes the
corrupted prompt into a coherent problem and reports confidence in the solution to \emph{that}
problem, so confidence should track internal consistency rather than input fidelity. We test this
\emph{behaviorally} with a \emph{restate-then-resolve} probe on the $n{=}400$ held-out noisy prompts---a
functional probe of what the model does with the corrupted input, not a causal manipulation of its hidden
state---run identically for
the DLM and the AR baseline. For each committed answer we record its confidence and correctness; for
every error we then ask the model to rewrite the corrupted prompt as a clean problem $R$ and re-solve
$R$ to obtain $A_R$. We measure reconstruction \emph{fidelity} (fraction of the clean problem's
numerals recovered in $R$), \emph{recovery} ($A_R$ equals gold), and \emph{coherent-misread} ($A_R$
reproduces the committed error), with $10^4$-resample bootstrap intervals (Table~\ref{tab:mech},
visualized as Figure~\ref{fig:mechanism} of the main paper). Two
facts emerge. \emph{(i) The confidence degradation is DLM-specific and complete:} every one of the DLM's
$232$ errors is high-confidence (mean $0.983$, statistically indistinguishable from $0.986$ on correct
answers), whereas the AR is confident on only $34\%$ of its errors at $\mathrm{conf}\,{\ge}\,0.9$ and
$2\%$ at ${\ge}\,0.95$ (mean error-confidence $0.887$, below its $0.901$ on correct answers).
\emph{(ii) The two families are indistinguishable on what they do with the corrupted input:} both
reconstruct the clean problem at similar fidelity ($0.738$ vs.\ $0.687$), reproduce the committed error
at similar rates ($0.272$ vs.\ $0.269$), and recover gold at similar rates ($0.233$ vs.\ $0.226$)---every
DLM$-$AR difference has a bootstrap CI spanning $0$. (Recovery and misread do not sum to one because
the remaining re-solves diverge from both, reflecting compounding noise in the two-step chain, e.g.\ an
occasional language switch in the restatement.) The DLM can reconstruct the true problem when
asked---the correct reading is accessible---yet its \emph{direct} answer is confidently wrong. The
families differ not in whether they repair the input but in whether their confidence registers that a
repair was needed: the AR's does, the DLM's does not. This behavioral signature is
\emph{consistent with} the ``consistency without fidelity'' mechanism, which we confirm directly below
with a causal intervention on the hidden state (Figure~\ref{fig:causal}); the readout exploits it by
scoring correctness from the representation rather than the saturated confidence.

\paragraph{A causal intervention on the hidden state.}
The restate-then-resolve probe is behavioral; we now intervene directly on the representation the
readout scores, to test whether its correctness axis is \emph{causally} wired into the model or merely a
passive correlate. Using the same frozen readout, we form its raw-space correctness direction
$\bm{d}=\bm{u}/\lVert\bm{u}\rVert_2^2$ and add $\alpha\bm{d}$ to each answer-token residual at layer $\ell$
through a forward hook---because the readout mean-pools the answer span, this shifts its logit by exactly
$\alpha$---measuring the effect in a teacher-forced pass over the fixed prompt--answer sequence and
sweeping $\alpha\in[-4,4]$ on the $n{=}200$ held-out noisy prompts against norm-matched random control
directions (Figure~\ref{fig:causal}). Three findings emerge.
\emph{(i) The handle is calibrated} (manipulation check): by construction $\bm{u}^{\top}\bm{d}{=}1$, and
empirically the intervention moves the readout logit by precisely $\alpha$ (slope $1.00$, $95\%$ CI
$[1.00,1.00]$), confirming the hook fires at the intended layer and the pooled readout responds as
designed; this fixes the scale of $\alpha$ (one unit $=$ one logit of readout) for the causal test that
follows. \emph{(ii) It causally and specifically shifts the model's own answer:} steering monotonically
changes the log-probability the model assigns to its committed answer ($-0.0016$ per unit $\alpha$, paired
bootstrap $P{<}0.001$), whereas norm-matched random directions of equal magnitude move it far less and in
the \emph{opposite} direction (paired readout$-$control difference $-0.0019$, $P{<}0.001$)---so the signed
de-commitment is carried by the correctness direction itself, not by generic perturbation of the residual
stream. \emph{(iii) The sign is de-commitment, not correction:} pushing the representation \emph{toward}
the readout's ``correct'' side \emph{lowers} the likelihood of the answer the model already committed to,
and does so similarly whether that answer was in fact correct ($-0.0014$) or incorrect ($-0.0017$;
difference $+0.0003$, $P{=}0.30$). The readout axis is thus locally \emph{opposed} to committed-answer
likelihood---consistent with ``consistency without fidelity,'' in which the model's endorsement tracks
structural coherence rather than correctness. Importantly, the intervention is a handle on \emph{confidence},
not a controller of \emph{correctness}: it does not raise the gold-answer likelihood ($+0.0002$ per
$\alpha$, $P{=}0.60$). A stronger, confound-free confirmation comes from re-running the \emph{full
diffusion generation} under the same intervention on a separate $n{=}100$ set (the gold answer is never
visible): the readout handle remains exact \emph{during decoding} (slope $1.00$), yet re-decoded accuracy
does not move (LCR slope $+0.004$ per $\alpha$, $P{=}0.33$) and is indistinguishable from the random
control (specificity $P{=}0.79$). This is precisely why the effective remedy is a training-free
\emph{readout} for
selective prediction rather than a decoding-time steering intervention: the corrupted reading is fixed
upstream in prompt encoding (as argued in the main paper), so the answer representation faithfully \emph{carries}
the correctness signal---which the readout recovers---but steering it does not, by itself, raise the
likelihood of the correct answer, consistent with the baseline's finding that prompt- and decoding-level
patching fails.

\begin{table}[t]
\centering
\small
\setlength{\tabcolsep}{4pt}
\resizebox{\columnwidth}{!}{%
\begin{tabular}{lccccc}
\toprule
Model & \multicolumn{2}{c}{Conf.\ on errors} & Fidelity & Recovery & Misread \\
\cmidrule(lr){2-3}
      & mean & $\%{\ge}.95$ & & {\footnotesize $A_R{=}$g} & {\footnotesize $A_R{=}A$} \\
\midrule
LLaDA-8B (DLM)     & $0.983$ & $100$ & $0.738$ & $0.233$ & $0.272$ \\
LLaMA-3-8B (AR) & $0.887$ & $2$   & $0.687$ & $0.226$ & $0.269$ \\
\bottomrule
\end{tabular}}
\caption{Restate-then-resolve mechanism probe (GSM8K, transposition $\eta{=}0.15$, $n{=}400$ held-out noisy prompts). ``Conf.\ on errors'' covers all committed errors (DLM $232$, AR $274$); fidelity, recovery and misread cover confident errors ($\mathrm{conf}{\ge}0.9$: DLM $232$, AR $93$), each a $10^4$-bootstrap estimate. The DLM is confident on every error, yet both models reconstruct the clean problem, recover gold ($A_R{=}$g) and reproduce their committed error ($A_R{=}A$) at indistinguishable rates (all pairwise $\Delta$ CIs include $0$).}
\label{tab:mech}
\end{table}

\subsection{How far does one frozen readout reach?}
The generalization experiments above re-fit the readout per condition. A stronger question is
how far a \emph{single} frozen readout reaches: we take one probe---fit once on the GSM8K
clean$+$transposition calibration split (mean-pooled $\ell{=}20$, $C{=}10^{-3}$)---and apply it
\emph{with no refit} to eight target test splits spanning three axes
(Table~\ref{tab:frozenreach}). The frozen probe transfers across \emph{noise family} (keyboard
$+0.126$, insertion $+0.108$, both significant; deletion $+0.035$, n.s.---the same
length-alignment boundary as before) and across \emph{severity} (a probe fit at $\eta{=}0.15$ still
adds $+0.138$, $+0.133$, and $+0.069$ at $\eta{=}0.05,0.10,0.30$): it never sees these conditions
yet recovers most of the in-domain gain. The boundary is \emph{task}: the same frozen
GSM8K probe adds only $+0.021$ (n.s.) on ARC-Challenge, even though a probe \emph{re-fit} on ARC
recovers a significant $+0.105$ (the ARC subsection above). The correctness direction is
therefore \textbf{noise- and severity-invariant but task-specific}: one probe covers the
corruption families and intensities a deployed model meets on a given task, but a new task needs
its own one-time calibration fit. This localizes---rather than inflates---the method's reach.

\begin{table}[t]
\centering
\small
\setlength{\tabcolsep}{4pt}
\resizebox{\columnwidth}{!}{%
\begin{tabular}{llccc}
\toprule
Target split (one frozen probe) & Axis & Base & Frozen LCR & $\Delta$ (P) \\
\midrule
GSM8K transposition (in-domain) & ---      & $0.575$ & \best{$\mathbf{0.720}$} & $+0.145$ ($1.00$) \\
GSM8K keyboard                  & noise    & $0.613$ & \best{$\mathbf{0.740}$} & $+0.126$ ($0.997$) \\
GSM8K insertion                 & noise    & $0.652$ & \best{$\mathbf{0.759}$} & $+0.108$ ($0.995$) \\
GSM8K deletion                  & noise    & $0.589$ & $0.624$ & $+0.035$ ($0.778$) \\
GSM8K $\eta{=}0.05$                & severity & $0.627$ & \best{$\mathbf{0.766}$} & $+0.138$ ($1.00$) \\
GSM8K $\eta{=}0.10$                & severity & $0.614$ & \best{$\mathbf{0.747}$} & $+0.133$ ($1.00$) \\
GSM8K $\eta{=}0.30$                & severity & $0.671$ & \best{$\mathbf{0.740}$} & $+0.069$ ($0.96$) \\
ARC-Challenge                   & task     & $0.483$ & $0.503$ & $+0.021$ ($0.80$) \\
\bottomrule
\end{tabular}}
\caption{Reach of \emph{one} frozen readout (GSM8K/transposition anchor, mean-pooled $\ell{=}20$, $C{=}10^{-3}$) applied with no refit to eight target splits (selective AUROC; $n{=}400$ in-domain, severity and ARC, $n{=}200$ for the held-out noise families). It transfers across noise family and severity but not across task, where a re-fit probe instead gains $+0.105$: the correctness direction is noise- and severity-invariant but task-specific. $P$ is the paired-bootstrap probability $\Delta{>}0$; bold marks significant gains over base confidence.}
\label{tab:frozenreach}
\end{table}

\subsection{Corruption severity}
\label{sec:sup-severity}
\begin{figure}[t]
\centering
\includegraphics[width=\columnwidth]{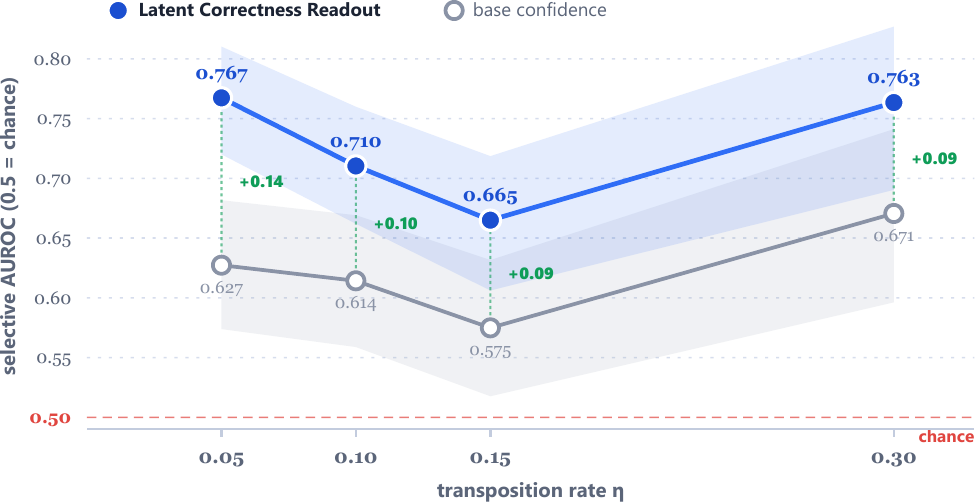}
\caption{Ranking ability against corruption severity on GSM8K. The readout stays above base confidence at every severity and the curves do not converge, so the gain is not an artifact of one operating point.}
\label{fig:severity}
\end{figure}

The main paper runs at a single severity, $\eta=0.15$, so we check that the ranking gain is not
an artifact of that operating point. Figure~\ref{fig:severity} sweeps transposition severity on
GSM8K from $\eta=0.05$ to $\eta=0.30$, re-fitting the readout at each level under the usual
protocol. LCR stays above base confidence throughout---$+0.139$, $+0.096$, $+0.145$ and $+0.093$ at
$\eta=0.05$, $0.10$, $0.15$ and $0.30$---and every bootstrap interval excludes zero at
$P\ge0.98$. Accuracy falls steadily
with severity while the ranking gain does not, and the two curves do not converge as corruption
grows. The frozen-probe rows of Table~\ref{tab:frozenreach} make the same point without refitting:
a probe fit once at $\eta=0.15$ still adds $+0.138$, $+0.133$ and $+0.069$ at $\eta=0.05$, $0.10$
and $0.30$.

\subsection{Real-world noise: human typos}
Our headline corruption is synthetic character transposition. To check that the confidence readout
reflects the phenomenon rather than that particular process, we repeat the noisy-only protocol on
\emph{real} human misspellings, replacing whole words with attested misspellings from a
machine-readable corpus (Wikipedia's list of common misspellings; $3{,}212$ single-word entries).
This corruption differs from transposition in two ways we state plainly: it is drawn from real human
errors, and---because the corpus contains no numerals---it leaves the quantities in each problem
intact. At a per-word rate of $0.6$ it changes $15.9\%$ of words (comparable to the $\eta{=}0.15$
transposition headline) and touches every test prompt, while preserving the numerals in all $400$.

Two consequences follow, and we report both rather than only the favourable one
(Table~\ref{tab:realtypo}). First, because the numbers survive, the task stays largely solvable
(LLaDA\ test accuracy $0.765$ versus $0.37$ under transposition), so the base-confidence degradation is
milder: held-out selective AUROC falls from $0.666$ (clean) to $0.633$ under real typos, against
$0.575$ under transposition. Second---and this is the point of the check---the readout still recovers
a significant amount of ranking ability, $+0.148$ (CI $[+0.074,+0.221]$, $P{=}1.000$), lifting
selective AUROC to $0.781$, a gain at least as large as the $+0.090$ under synthetic noise. The
readout is therefore not an artifact of character transposition; its benefit persists under a
non-synthetic input distribution, while the depth of the base degradation tracks how much the
corruption actually damages the problem.

\begin{table}[t]
\centering
\small
\setlength{\tabcolsep}{4pt}
\resizebox{\columnwidth}{!}{%
\begin{tabular}{lccc}
\toprule
Noise process & Acc. & Base (clean$\to$noisy) & Readout ($\Delta$, $P$) \\
\midrule
Transposition ($\eta{=}0.15$, synthetic) & $0.37$  & $0.666{\to}0.575$ & \best{$\mathbf{0.665}$ (${+}0.090$, $0.99$)} \\
Real typos (human, numerals kept)     & $0.765$ & $0.666{\to}0.633$ & \best{$\mathbf{0.781}$ (${+}0.148$, $1.00$)} \\
\bottomrule
\end{tabular}}
\caption{The confidence readout replicates under real human typos (same noisy-only protocol, LLaDA, GSM8K, $n_{\mathrm{test}}{=}400$). Real misspellings preserve the numerals, so the task stays solvable and base degradation is milder than under synthetic transposition; the readout's recovery nonetheless replicates and is at least as large. Bold marks the better ranker under either noise process. Source: \texttt{ncc\_realtypo/realtypo\_metrics.json}.}
\label{tab:realtypo}
\end{table}

\subsection{Ablations and analysis}
\begin{figure}[t]
\centering
\includegraphics[width=\columnwidth]{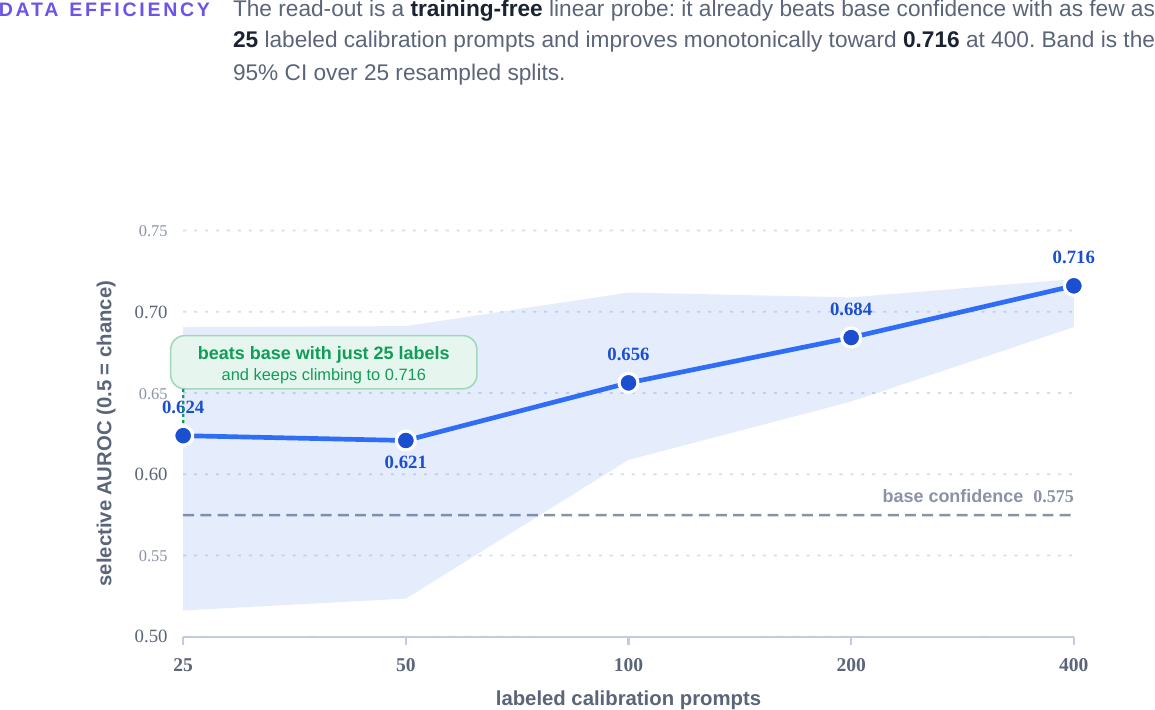}
\caption{Data efficiency of the readout. Selective AUROC against the labeled calibration budget $n_{\mathrm{fit}}$ (log axis), mean over $R{=}25$ resampled splits with a $95\%$ CI band. The readout beats base confidence ($0.575$, dashed) at every budget, is flat between $n_{\mathrm{fit}}{=}25$ and $50$ ($0.624$, $0.621$), and rises to $0.716$ at $400$.}
\label{fig:learning}
\end{figure}

Five analyses support the readout. \emph{(i) The routing negative is real, not
underpowered.} At $n{=}400$, NCC-Train exceeds base by $+0.057$ selective AUROC (CI
$[-0.015,0.134]$, n.s.) and differs from consistency by $-0.005$ (CI $[-0.072,0.061]$): the
probe-gated penalty is indistinguishable from plain consistency, so the readout's gain
does not come from the corruption signal. \emph{(ii) Layer and pooling.} Selective AUROC is
highest for mean-pooled states at intermediate-to-late layers (L16--L20); CV on the
calibration split selects this region without peeking at the test set. \emph{(iii)
Abstention.} Ranking by the readout and deferring the lowest-scoring answers cuts noisy
error from $0.58$ to $0.48$ at $50\%$ coverage and lowers the area under the risk--coverage
curve from $0.57$ (base) to $0.46$, making the readout directly actionable for selective
prediction. \emph{(iv) The degradation is not an artifact of corrupted labels.} Character
transposition does not protect numerals, so a corrupted GSM8K prompt could in principle alter
the arithmetic and invalidate its clean gold answer, manufacturing a spurious drop in accuracy
and confidence quality. We therefore split the noisy test set by whether every numeric token
survives corruption. On the $46\%$ of prompts that are \emph{numeral-preserving}---where the
clean gold provably still applies---base confidence is if anything \emph{more} degraded
(selective AUROC $0.550$) and the readout recovers it \emph{more} strongly ($0.732$, a
$+0.182$ gain) than on the full test set; on numeral-altered prompts the same pattern is weaker
but present (base $0.564$, readout $0.664$, $+0.099$). The representation--confidence gap and
its readout remedy are therefore not driven by label noise from corrupted problems---they are
strongest where the gold answer is guaranteed intact. Because the numeral test is a
surface heuristic, we add an independent, model-based criterion: a held-out instruction model
(LLaMA-3-8B, distinct from the confidence signal) reads each original and corrupted problem and
judges, through the typos, whether the numbers and question are unchanged. This judge is stricter
than the numeral test and only weakly correlated with it ($28\%$ vs.\ $46\%$ of prompts judged
preserving; Cohen $\kappa{=}0.08$), reflecting that the two operationalize preservation
differently and that neither replaces human annotation. Even so, the readout gain is essentially
unchanged on the judge-preserving subset ($n{=}114$, $0.542{\to}0.691$, $+0.149$), and accuracy
there is if anything lower than on the full set ($0.333$ vs.\ $0.372$): on inputs an independent
model reads as answer-preserving, the model's confidence still degrades and the readout still recovers the
ordering. The effect is thus stable across two different notions of meaning preservation.
\emph{(v) The gain is stable
across resampling and calibration budget.} Over $R{=}25$ random re-partitions of the noisy
pool into $200$-calibration / $400$-test (the clean calibration held fixed), the readout
averages selective AUROC $0.748$ (CI $[0.719,0.779]$) against base $0.589$ (CI
$[0.557,0.621]$)---a $+0.159$ gap (CI $[+0.114,+0.190]$) that is positive on \emph{all} $25$
splits---bracketing the fixed-seed combined-fit value of $0.720$. A learning curve over the labeled
calibration budget clears base confidence at every budget and rises with it after an initial
plateau (Figure~\ref{fig:learning})---$0.624$ ($n_{\mathrm{fit}}{=}25$), $0.621$ ($50$), $0.656$
($100$), $0.684$ ($200$), $0.716$ ($400$)---so even a $25$-example budget already beats base, and
the method is not relying on an unusually favorable split.

\section{Additional supporting results}

\subsection{The accuracy axis is largely recovered by a simple baseline}
Table~\ref{tab:accuracy} reports accuracy under noise on the testbed cell. The
consistency baseline lifts noisy accuracy from $0.370$ to $0.490$ (a $+0.120$ paired
gain, significant). The three diffusion-native elaborations do not improve on it: a
timestep-stratified consistency schedule ties it ($+0.010$, n.s.); a hidden-state
corruption-invariance penalty ties it ($-0.020$, n.s.); and a multi-step rollout
consistency is \emph{significantly worse} ($-0.170$; McNemar $p{=}0.002$), because the
free-running student chases its own compounding errors and never learns to answer. These
negative results are, we argue, informative: within this fixed $n{=}100$ testbed cell the ordering is what matters---our
diffusion-native elaborations do not beat plain consistency, and one is significantly
worse---so we read it as a \emph{relative} result that motivates reallocating effort to
reliability, not as a claim of a universal accuracy ceiling. The larger $n{=}400$ reliability
experiments below inherit consistency's accuracy ($0.41$) and study the confidence axis on
top of it.

\begin{table}[t]
\centering
\small
\setlength{\tabcolsep}{4pt}
\resizebox{\columnwidth}{!}{%
\begin{tabular}{lccc}
\toprule
Method & Noisy acc. & $\Delta$ vs.\ base & vs.\ baseline \\
\midrule
Base (no tuning)            & $0.370$ & --      & --  \\
Consistency (baseline)      & \sbest{$\mathit{0.490}$} & $+0.120$$^\ast$ & --  \\
\;+ stratified schedule     & \best{$\mathbf{0.500}$} & $+0.130$ & $+0.010$ n.s. \\
\;+ invariance penalty      & $0.470$ & $+0.100$ & $-0.020$ n.s. \\
\;+ rollout consistency     & $0.320$ & $-0.050$ & $-0.170^\ast$ \\
\bottomrule
\end{tabular}}
\caption{Accuracy under noise (GSM8K, transposition $\eta{=}0.15$). A simple consistency objective (italics) recovers most lost accuracy; three diffusion-native elaborations do not beat it, and the numerically best variant (bold) is within noise of the baseline while one is significantly worse. $^\ast$: significant under paired bootstrap / McNemar.}
\label{tab:accuracy}
\end{table}

\subsection{The gain requires the hidden state, not a cheaper signal}
Could a much cheaper single-pass signal match the readout at the same supervised budget? We
fit ten same-budget competitors on the identical clean$+$noisy calibration split and score
them on the same $n{=}400$ noisy test set (Table~\ref{tab:samebudget}): the raw confidence; its
mean- and minimum-token variants; the mean and maximum per-token predictive entropy; the
decoded answer length; a supervised Platt rescaling of the confidence; a supervised
non-monotone histogram binning of it; the score of a supervised corruption detector read off
the same hidden states; and a supervised two-feature confidence-plus-corruption bundle. None
clears $0.59$ selective AUROC. The obvious uncertainty proxies a practitioner would try
first---least-confident token, per-token entropy, answer length---are no better than the
aggregate confidence, all within noise of chance. Monotone
rescaling cannot move the ranking at all ($0.575$, by construction); letting the recalibrator be
non-monotone \emph{lowers} it ($0.562$), i.e.\ the confidence scalar carries too little
correctness information to re-rank usefully. Most informative is the corruption detector: on these
same hidden states it separates clean from corrupted inputs almost perfectly (clean-vs-corrupt
AUROC $0.986$), yet as a \emph{correctness} ranker it is barely above chance ($0.548$), and
adding it to confidence buys only $+0.006$ ($0.581$). The full $4096$-dimensional readout
reaches $0.720$, a $+0.139$ gain over that confidence-plus-corruption bundle (paired bootstrap CI
$[+0.076,+0.199]$, $P{=}1.000$). The correctness signal is thus broadly \emph{distributed} in
the answer-token representation: it is not the model's confidence, not a one-dimensional
``how corrupted is the input'' axis, and not their linear combination---consistent with our
thesis that input-level corruption and answer-level correctness are different variables.

\begin{table}[t]
\centering
\small
\setlength{\tabcolsep}{4pt}
\resizebox{\columnwidth}{!}{%
\begin{tabular}{lcc}
\toprule
Single-pass signal (same budget) & Sel.\ AUROC & $95\%$ CI \\
\midrule
Raw confidence (base)                         & $0.575$ & $[0.517,0.633]$ \\
Mean-token confidence (sup., $1$-d)           & \sbest{$\mathit{0.585}$} & $[0.527,0.643]$ \\
Min-token confidence (sup., $1$-d)            & $0.517$ & $[0.457,0.577]$ \\
Mean-token entropy (sup., $1$-d)              & $0.584$ & $[0.526,0.640]$ \\
Max-token entropy (sup., $1$-d)               & $0.503$ & $[0.442,0.563]$ \\
Answer length (sup., $1$-d)                   & $0.535$ & $[0.474,0.594]$ \\
Platt confidence (sup., monotone)             & $0.575$ & $[0.517,0.633]$ \\
Hist-bin confidence (sup., non-mono.)         & $0.562$ & $[0.507,0.620]$ \\
Corruption score (sup., $1$-d)                & $0.548$ & $[0.490,0.607]$ \\
Conf.\ $+$ corruption (sup., $2$-d)           & $0.581$ & $[0.523,0.638]$ \\
\midrule
\textbf{LCR\ hidden states (sup., $4096$-d)} & \best{$\mathbf{0.720}$} & $[0.668,0.771]$ \\
\bottomrule
\end{tabular}}
\caption{Same-budget single-pass baselines (GSM8K, transposition $\eta{=}0.15$, $n_{\mathrm{test}}{=}400$, all fit on the same clean and noisy calibration split). No cheap scalar clears $0.59$ selective AUROC (italics mark the strongest); only the full hidden-state readout does (bold, $+0.139$ over the confidence-plus-corruption bundle, CI $[+0.076,+0.199]$). The corruption detector flags corrupted inputs almost perfectly yet ranks correctness at chance: corruption and correctness are distinct variables.}
\label{tab:samebudget}
\end{table}

\subsection{Combining the readout with cheap sampling}
Self-consistency ranks correctness better than the readout on GSM8K ($0.842$ vs.\ $0.665$
selective AUROC) but costs $k$ additional full generations per input, whereas the readout is a
single extra hidden-state read. It is therefore worth asking whether the free readout and a small
sampling budget can be combined so that, when only a few samples are affordable, the pair improves
on sampling alone. We test the simplest parameter-free fusion: for a budget of $k$ extra generations
we form the self-consistency signal from $k$ samples, form the readout score, and combine them by
averaging their ranks; we compare this against spending the same $k$ generations on self-consistency
alone. Significance is a paired bootstrap over prompts that redraws the $k$-sample subset on each
resample.

The free readout is worth roughly one extra generation in the low-budget regime. At $k{=}1$ the
fusion reaches $0.754$ against $0.711$ for one-sample self-consistency (paired gap $+0.043$,
$95\%$ CI $[+0.004,+0.082]$, $P(\text{fusion}{>}\text{sampling}){=}0.99$; Table~\ref{tab:cheaphybrid}).
The advantage is modest and confined to the smallest budget: at $k{=}2$ the gap is already only $+0.008$ ($P{=}0.68$), by $k{=}3$ sampling alone is ahead,
and by $k{=}10$ self-consistency ($0.842$) is well ahead. The readout thus helps where sampling is
least affordable and is dominated once several samples can be drawn; it complements rather than
replaces sampling.

\begin{table}[t]
\centering
\small
\setlength{\tabcolsep}{6pt}
\begin{tabular}{cccc}
\toprule
Budget $k$ & sampling only & readout $+$ sampling & $\Delta$ \\
\midrule
$1$  & $0.711$ & \best{$\mathbf{0.754}$} & $+0.043$ \\
$2$  & $0.771$ & \best{$\mathbf{0.779}$} & $+0.008$ \\
$3$  & \best{$\mathbf{0.799}$} & $0.791$ & $-0.008$ \\
$5$  & \best{$\mathbf{0.823}$} & $0.801$ & $-0.021$ \\
$10$ & \best{$\mathbf{0.842}$} & $0.809$ & $-0.033$ \\
\bottomrule
\end{tabular}
\caption{Combining the free readout with cheap sampling. Selective AUROC on GSM8K deployment noise ($n_{\text{test}}{=}400$) for $k$-sample self-consistency alone against a parameter-free rank-average of the readout with the same $k$ samples. Bold marks the better option at each budget: the fusion helps at $k{=}1$ ($P{=}0.99$), is indistinguishable at $k{=}2$ ($P{=}0.68$), and falls behind thereafter. For reference the readout alone scores $0.665$ and base confidence $0.575$.}
\label{tab:cheaphybrid}
\end{table}

\section{Robustness of the predictive diagnostic}
The main-paper diagnostic (Table~\ref{tab:predict}) sorts eleven tasks
into five the readout helps and six it does not, using a fit-only margin and a leave-one-task-out
threshold. Three independent checks confirm that this split is not an artifact of a single
noise draw, a single calibration sample, or the particular AUROC metric.

\subsection{Stability across input-noise realizations}
The headline gains fix one random seed for the input corruption. To test sensitivity to the noise
draw we re-run the two clearest beneficiaries, GSM8K and ARC-Challenge, under three further seeds
(the seed controls which characters are transposed; seed $42$ is the original headline run), re-fitting
the readout per seed with the identical CV protocol (Table~\ref{tab:multiseed}). ARC is uniformly
robust: the readout beats base confidence on \emph{all four} seeds ($\Delta{=}+0.078{\pm}0.036$,
every seed positive; a one-sample $t$-test across the four seed-level gains rejects zero, $t_3{=}4.31$, $p{=}0.02$, $95\%$ CI $[+0.021,+0.136]$). GSM8K is positive on three of four seeds
($\Delta{=}+0.073{\pm}0.076$); the one negative draw (seed $44$) is not an unusually harsh corruption, since its base AUROC of $0.626$ sits \emph{above} the four-seed base mean of $0.600$. What distinguishes it is model selection: cross-validation chose last-token pooling at $C{=}0.1$ on that seed, whereas all three positive seeds chose mean pooling at $C{=}0.001$ to $0.003$, a regularization strength some thirty times tighter. The failure is therefore in the selection step rather than in the noise, which matters practically because a deployer cannot audit that step without labels. The other three seeds range $+0.090$ to $+0.137$. Because of that spread the four-seed GSM8K mean is not significantly above zero at this sample size ($t_3{=}1.92$, $p{=}0.15$, $95\%$ CI $[-0.048,+0.195]$); we report it as positive-on-average but seed-sensitive and rest the cross-seed significance claim on ARC. The diagnostic's beneficiary call
is stable to the noise seed for both tasks---the calibration-set margin selects GSM8K and ARC at
every seed; it is the realized GSM8K \emph{gain} that varies with the noise draw, as expected of a
task whose base confidence is already comparatively strong.

\begin{table}[t]
\centering
\small
\setlength{\tabcolsep}{5pt}
\resizebox{\columnwidth}{!}{%
\begin{tabular}{lccccc}
\toprule
Task & seed $42$ & $43$ & $44$ & $45$ & mean$\pm$sd \\
\midrule
GSM8K         & $+0.090$ & $+0.137$ & $-0.037$ & $+0.104$ & $+0.073{\pm}0.076$ \\
ARC-Challenge & $+0.104$ & $+0.089$ & $+0.025$ & $+0.096$ & $+0.078{\pm}0.036$ \\
\bottomrule
\end{tabular}}
\caption{The readout gain across input-noise seeds. Per-seed held-out gain $\Delta$ (LCR\ $-$ base) under transposition noise ($\eta{=}0.15$, $n_{\mathrm{test}}{=}400$), re-fit independently at each seed. ARC is positive on all four seeds with a cross-seed mean above zero ($t_3{=}4.31$, $p{=}0.02$); GSM8K is positive on three of four (seed $44$ is a model-selection failure rather than a harsh corruption) but its mean is not significant at this sample size ($t_3{=}1.92$, $p{=}0.15$). Seed $42$ is the headline run.}
\label{tab:multiseed}
\end{table}

\subsection{Stability across the calibration sample}
The margin and the deployed readout are estimated on one calibration split. To check that the
help/null verdict does not hinge on that particular sample, we bootstrap-resample the calibration
set ($B{=}1000$), re-fit the deployed readout (fixed CV-selected layer and regularizer) on each
resample, and re-evaluate on the fixed noisy test set. The verdict is invariant: on \emph{every}
one of the five beneficiaries the refit gain stays positive with high probability---ARC
$+0.087$ ($P{=}1.00$), GSM8K $+0.070$ ($P{=}1.00$), SciQ $+0.083$ ($P{=}1.00$), TriviaQA $+0.074$
($P{=}0.99$), SVAMP $+0.186$ ($P{=}1.00$)---and on \emph{every} one of the six nulls it does not,
three of them turning significantly negative (WinoGrande $-0.067$, $P{<}0.01$; Social IQa $-0.033$,
$P{=}0.02$; HellaSwag $-0.027$, $P{=}0.04$; CommonsenseQA, PIQA, OpenBookQA null). Which tasks
benefit is therefore a property of the task, not of the calibration draw, and the three tasks the
diagnostic most confidently withholds are those a refit would have harmed.

\subsection{A deployment-facing metric agrees}
Selective AUROC ranks answers; a deployer ultimately cares about the accuracy retained after
abstaining on the least-trusted answers. We therefore score each task by selective accuracy at
$50\%$ coverage and by the area under the risk--coverage curve (AURC, lower is better), comparing the
readout to base confidence (Table~\ref{tab:selpred}). The deployment metric reproduces the split:
all five beneficiaries gain selective accuracy and lower AURC---SVAMP $0.540{\to}0.690$
(AURC $0.446{\to}0.285$), ARC $0.580{\to}0.680$, SciQ $0.670{\to}0.725$, GSM8K $0.425{\to}0.505$,
TriviaQA $0.345{\to}0.405$, all with $P(\text{LCR}\ \text{better}){\ge}0.82$---while no null improves
and WinoGrande, the task the readout hurts, gets worse ($0.590{\to}0.555$, AURC $0.387{\to}0.453$).
The readout's advantage is thus not a quirk of the ranking metric: it delivers the operationally
relevant quantity---more correct answers retained at fixed coverage---on precisely the tasks the
fit-only diagnostic green-lights.

\begin{table}[t]
\centering
\small
\setlength{\tabcolsep}{4pt}
\resizebox{\columnwidth}{!}{%
\begin{tabular}{lccc}
\toprule
Task & sel-acc @$50\%$ & AURC & $P(\text{LCR}{>}\text{base})$ \\
 & (base$\to$LCR) & (base$\to$LCR) & \\
\midrule
ARC-Challenge & $0.580\!\to\!0.680$ & $0.368\!\to\!0.323$ & \best{$\mathbf{0.91}$} \\
GSM8K         & $0.425\!\to\!0.505$ & $0.568\!\to\!0.523$ & \best{$\mathbf{0.93}$} \\
SciQ          & $0.670\!\to\!0.725$ & $0.317\!\to\!0.275$ & \best{$\mathbf{0.92}$} \\
TriviaQA      & $0.345\!\to\!0.405$ & $0.646\!\to\!0.617$ & \best{$\mathbf{0.82}$} \\
SVAMP         & $0.540\!\to\!0.690$ & $0.446\!\to\!0.285$ & \best{$\mathbf{1.00}$} \\
\midrule
WinoGrande    & $0.590\!\to\!0.555$ & $0.387\!\to\!0.453$ & $0.02$ \\
PIQA          & $0.680\!\to\!0.675$ & $0.297\!\to\!0.329$ & $0.14$ \\
CommonsenseQA & $0.545\!\to\!0.510$ & $0.445\!\to\!0.486$ & $0.12$ \\
Social IQa    & $0.575\!\to\!0.525$ & $0.422\!\to\!0.438$ & $0.31$ \\
OpenBookQA    & $0.475\!\to\!0.475$ & $0.486\!\to\!0.494$ & $0.39$ \\
HellaSwag     & $0.455\!\to\!0.455$ & $0.524\!\to\!0.553$ & $0.18$ \\
\bottomrule
\end{tabular}}
\caption{A deployment-facing metric reproduces the help/null split. Selective accuracy at $50\%$ coverage (higher is better) and AURC (lower is better) under deployment noise, base confidence against the readout; $P$ is the paired-bootstrap probability the readout is better, bold where it wins. All five beneficiaries improve on both metrics, no null task does, and WinoGrande worsens.}
\label{tab:selpred}
\end{table}

\subsection{An end-to-end abstention gate}
The selective metrics above rank answers; a deployer needs a gate that is fixed on calibration data
and then applied online. We freeze the readout (layer $16$, mean-pooled, $C{=}0.003$) on the $200$
noisy calibration prompts, storing the standardizer, the linear weights, and per-coverage thresholds
as a small serialized artifact. At serve time each new noisy input contributes one hidden-state read,
the frozen probe scores it, and the gate answers the top-coverage fraction of the stream.

On the $400$ held-out noisy prompts the frozen gate retains more accuracy than a base-confidence gate
at every coverage (Table~\ref{tab:gatedemo}): at $50\%$ coverage the readout gate keeps $0.505$
against $0.425$ for the base-confidence gate and $0.372$ for answering every input, and the ordering
holds at $30\%$ ($0.533$ vs.\ $0.442$) and $70\%$ ($0.443$ vs.\ $0.404$). These values reproduce the
GSM8K row of Table~\ref{tab:selpred} from a frozen, serialized probe rather than a re-fit. As an
execution check we also ran the frozen probe on the diffusion model over $40$ further held-out noisy
prompts end to end---the model generates, the probe consumes the live hidden states, and the gate
fires online---retaining $0.536$ against $0.500$ for answering all. At this small stream the
calibration threshold overshoots the intended coverage ($0.70$ actual vs.\ $0.50$ target), so the
coverage-controlled numbers are those from the $n{=}400$ validation; the live run establishes only
that the frozen probe drives a working gate outside the fitting pipeline.

\begin{table}[t]
\centering
\small
\setlength{\tabcolsep}{6pt}
\begin{tabular}{cccc}
\toprule
Coverage & answer all & base-conf.\ gate & readout gate \\
\midrule
$30\%$ & $0.372$ & $0.442$ & \best{$\mathbf{0.533}$} \\
$50\%$ & $0.372$ & $0.425$ & \best{$\mathbf{0.505}$} \\
$70\%$ & $0.372$ & $0.404$ & \best{$\mathbf{0.443}$} \\
\bottomrule
\end{tabular}
\caption{A frozen readout drives a working abstention gate. Retained accuracy on $400$ held-out noisy GSM8K prompts when the gate answers the top-coverage fraction by score, for answering everything, a base-confidence gate and the frozen readout. The readout gate retains the most accuracy at every coverage, and the $50\%$ column matches the GSM8K row of Table~\ref{tab:selpred}.}
\label{tab:gatedemo}
\end{table}

\subsection{The calibration gain is not a binning artifact}
The headline calibration improvement---base $ECE\;0.61$ to readout $ECE\;0.12$ on GSM8K under
deployment noise---is reported with $15$ equal-width bins. Because base confidence is saturated
(mean $0.98$), its probability mass sits in the top bin, a regime where equal-width ECE\ can in
principle be sensitive to the binning choice. We therefore recompute both ECE\ values under two
schemes---equal-width and equal-count (quantile) bins---at $10$, $15$, and $20$ bins. The base
ECE\ is identical ($0.610$) in all six settings---with saturated confidence
$ECE\approx|\overline{c}-\mathrm{acc}|$ irrespective of the bins---and the readout ECE\ stays in
$[0.115,0.124]$, so the base-to-readout reduction lies in a narrow band $[0.486,0.495]$ across every
scheme and bin count. The calibration gain is therefore a property of the scores, not of the
binning. 

\section{Limitations}
\label{sec:limits}

Several limitations qualify the result, and they fall into three groups.

The first concerns what the readout achieves. It is a modest ranker: its selective AUROC remains well short of what sampling achieves at any budget, and even a single extra generation ranks better. Its case rests on adding no generation at all, and on composing with sampling when the budget is small. Absolute selective performance under noise also stays far from what these systems would need in a setting where abstention genuinely matters.

The second concerns stability. The gain across independent noise draws is significant on ARC but only directional on GSM8K. The one negative draw traces to an unstable model-selection step, not to the corruption itself, which is a failure mode a deployer cannot detect without labels. The benefit is task-dependent, so the calibration-set screening rule belongs to the method and is not an optional extra, and that rule is itself validated on eleven tasks with a narrow margin. The method further assumes a labeled calibration set drawn from the deployment task, a stronger requirement than temperature scaling makes in practice, because the readout does not transfer across tasks the way a scalar does.

The third concerns the scope of our evidence. Our corruption model is programmatic and leaves numerals intact so that answers stay well defined, so the real-typo evaluation we report inherits that restriction and speaks to naturally occurring surface noise, not to adversarial or numeric corruption. The cross-family comparison is suggestive rather than exact, because confidence-ordered decoding selects the diffusion scores we read and no equivalent selection applies autoregressively; the behavioral evidence carries that claim. Finally, we study two diffusion models at the seven to eight billion parameter scale on English benchmarks, and whether the gap narrows with scale, with noise-aware instruction tuning, or in other languages remains open.


\end{document}